\documentclass[lettersize,journal]{IEEEtran}
\usepackage{algorithmic}
\usepackage{algorithm}
\usepackage{array}
\usepackage[caption=false,font=normalsize,labelfont=sf,textfont=sf]{subfig}
\usepackage{textcomp}
\usepackage{stfloats}
\usepackage{url}
\usepackage{verbatim}
\usepackage{graphicx}
\usepackage{cite}
\usepackage{flushend}
\usepackage{cite}
\usepackage{amsmath,amssymb,amsfonts}
\usepackage{xcolor}

\newcommand{\ignore}[1]{ }

\usepackage{multirow}
\usepackage{vcell}
\usepackage{mathtools}
\usepackage{multicol}
\usepackage{enumitem}

\usepackage[hidelinks, colorlinks=true, linkcolor=blue, citecolor=blue, anchorcolor=blue, filecolor=blue]{hyperref}

\usepackage{cleveref}

\usepackage[normalem]{ulem}
\usepackage{makecell}
\usepackage{tabularray}
\usepackage{colortbl}
\usepackage{hhline}

\usepackage{nccmath}

\newcommand{\blue}{}

\renewcommand{\blue}{\textcolor{blue}}

\usepackage{tikz,xcolor,hyperref}

\definecolor{lime}{HTML}{A6CE39}
\DeclareRobustCommand{\orcidicon}{%
	\begin{tikzpicture}
	\draw[lime, fill=lime] (0,0) 
	circle [radius=0.16] 
	node[white] {{\fontfamily{qag}\selectfont \tiny ID}};
	\draw[white, fill=white] (-0.0625,0.095) 
	circle [radius=0.007];
	\end{tikzpicture}
	\hspace{-2mm}
}

\foreach \x in {A, ..., Z}{%
	\expandafter\xdef\csname orcid\x\endcsname{\noexpand\href{https://orcid.org/\csname orcidauthor\x\endcsname}{\noexpand\orcidicon}}
}

\usepackage[prependcaption, textsize=footnotesize]{todonotes}

\usepackage{courier}

\begin{document}

\title{Sobol Sequence Optimization for Hardware-Efficient Vector Symbolic Architectures}

\author{       
        Sercan~Aygun \orcidA{}, \textit{Member, IEEE} and M.~Hassan~Najafi \orcidB{} \textit{Senior Member, IEEE}
\thanks{
   %\indent Manuscript received XXXXXXXXXX XX, 2023; accepted XXXXXXXXXX XX, 202X. Date of publication XXXXXXXXXX XX, 202X; date of current version XXXXXXXXXX XX, 2023. This article was recommended by Associate Editor XXXXXXXXX XXXXXXXXX. \textit{(Corresponding author: Sercan Aygun.)}
\\
  \indent The authors are %Sercan Aygun is 
  with the School of Computing and Informatics, University of Louisiana at Lafayette, Lafayette, LA, 70503, USA. E-mail: \{sercan.aygun, najafi\}@louisiana.edu.

   %\indent M. Hassan Najafi is with the School of Computing and Informatics, University of Louisiana at Lafayette, Lafayette, LA, 70503, USA (e-mail: najafi@louisiana.edu).

    %\indent Digital Object Identifier XXXXXXXXXXXXXXXXXXXXXXXXX
}

%\vspace{-15pt}

}

\markboth{}%
{Aygun and Najafi: Sobol Sequence Optimization for Hardware-Efficient Vector Symbolic Architectures}
%{Aygun and Najafi \MakeLowercase{\textit{et al.}}: A Sample Article Using IEEEtran.cls for IEEE Journals}


\maketitle

\begin{abstract}
Hyperdimensional computing (HDC) is an emerging computing paradigm with significant promise for efficient and robust learning. In HDC, objects are encoded with high-dimensional vector symbolic sequences called hypervectors. The quality of hypervectors, defined by their distribution and independence, directly impacts the performance of HDC systems. Despite a large body of work on the processing parts of HDC systems, little to no attention has been paid to data encoding and the quality of hypervectors. Most prior studies have generated hypervectors using inherent random functions, such as MATLAB's or Python's random function. This work introduces an optimization technique for generating hypervectors by employing quasi-random sequences. These sequences have recently demonstrated their effectiveness in achieving accurate and low-discrepancy data encoding in stochastic computing systems. The study outlines the optimization steps for utilizing Sobol sequences to produce high-quality hypervectors in HDC systems. An optimization algorithm is proposed to select the most suitable Sobol sequences for generating minimally correlated hypervectors, particularly in applications related to symbol-oriented architectures. The performance of the proposed technique is evaluated in comparison to two traditional approaches of generating hypervectors based on linear-feedback shift registers and MATLAB random function. The evaluation is conducted for two applications: (i) language and (ii) headline classification. Our experimental results demonstrate accuracy improvements of up to 10.79\%, depending on the vector size. Additionally, the proposed encoding hardware exhibits reduced energy consumption and a superior area-delay product.
\ignore{
Hyperdimensional computing (HDC) is an emerging computing paradigm that has shown significant promise for efficient and robust learning. In HDC, objects are encoded with high-dimensional vector symbolic sequences called hypervectors. The quality of hypervectors, defined by their distribution and independence, directly impacts the performance of HDC systems. 
%mimics important brain functionalities towards high-efficiency and noise-tolerant computation.
%Emerging technologies have recently been popular with paradigms such as stochastic computing (SC) and hyperdimensional computing (HDC).
Despite a large body of work on the processing parts of HDC systems, little to no attention has been paid to data encoding and the quality of hypervectors. Most prior works generate hypervectors using built-in random functions such as MATLAB or Python random functions.  
%Although many studies cover computing paradigms solitarily, blending different paradigms can yield fruitful results. 
%We propose a new computing avenue 
This work proposes an optimization technique for generating hypervectors %for HDC systems 
by utilizing %low-discrepancy (LD) 
quasi-random sequences. These sequences have been recently used for accurate and low-discrepancy encoding of data in stochastic computing systems. This study outlines optimization steps for Sobol sequences to generate high-quality hypervectors in HDC systems. %which are famous for providing high accuracy in SC. 
%A melting pot of these two concepts, we present a solution to the generation of uncorrelated vectors targeted for HDC systems. 
We propose an optimization algorithm to select the best Sobol sequences for generating minimally-correlated hypervectors for symbol-oriented language processing applications. 
%We decide on uncorrelated vectors with the proposed optimization algorithm applied to correlation distances from each Cartesian pair of hypervectors. 
%We use %recommend using 
%the stochastic correlation metric ($SCC$) to quantify independence between hypervectors. %for the first time in HDC systems. 
We evaluate the performance of the proposed technique compared to two traditional approaches of generating hypervectors based on linear-feedback shift registers and MATLAB random function for two different applications: (i) language and 
(ii) headline classification. Our experimental results show accuracy improvements of up to $10.79\%$ depending on the vector size. The proposed encoding hardware exhibits less energy consumption and better area-delay product.
%Considering the conventional random vector generation for the language classification problem, we prove the accuracy improvements up to $10.79\%$ depending on the vector size.
}
\end{abstract}


\section{Introduction}
\begin{figure}[t]
  \centering
  \includegraphics[width=\linewidth]{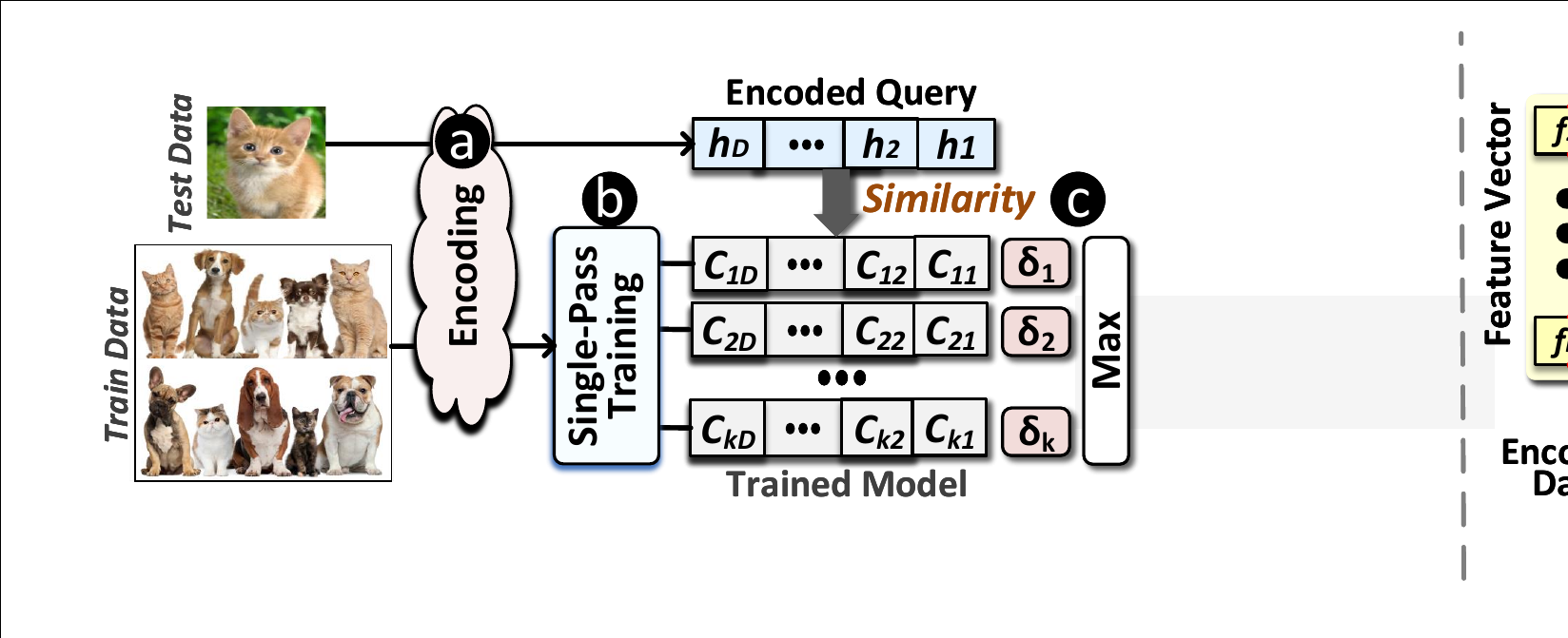}
  \caption{ Classification with hyperdimensional computing.}
  \label{Classification_HDC}
\end{figure}

\IEEEPARstart{H}{yperdimensional} computing (HDC)~\cite{aygun2023HDCencoding, kanerva2009hyperdimensional,imani2019framework} is a trending paradigm that mimics important brain functionalities toward high-efficiency and noise-tolerant computation. The paradigm has shown significant promise for efficient and robust learning \cite{10.1145/3453688.3461749}. HDC can transform data into knowledge at a very low cost and with better or comparable accuracy to state-of-the-art methods for diverse learning and cognitive applications~\cite{mitrokhin2019learning, RahimiGitHubBasedPaper}. 
The fundamental units of computation in HDC are high-dimensional vectors or “hypervectors” (consisting of $+1$s and $-1$s, or logic-$1$s and logic-$0$s) constructed from raw signals using an encoding procedure (Fig.~\ref{Classification_HDC}(a)).
During training, HDC superimposes together the encodings of signal values to create a composite representation of
a phenomenon of interest known as a “class hypervector” (Fig.~\ref{Classification_HDC}(b)). In inference, the nearest neighbor search identifies an appropriate class for the encoded query hypervector (Fig.~\ref{Classification_HDC}(c)).  Hypervectors have dimensionality, $D$, often in the orders of thousands of dimensions. A hypervector has a distributed, holographic representation %of data 
in which no dimension is more important than others. The hypervectors in an HDC system are generated to have almost \textit{zero} similarity. Previous works targeted \textit{near-orthogonal} hypervectors by generating random %binary 
hypervectors with approximately the same number of $+1$s and $-1$s~\cite{9995708, abbas-remat, najafabadi, SearcHD}. But the inherent randomness in these conventionally generated hypervectors can lead to poor performance, particularly with smaller $D$s. Low classification accuracy is likely in cases with poor distribution and undesired similarity %with non-uniform distribution of binary bits
%if there are some undesired similarities 
between hypervectors.  

%emulates the brain-like signals representing binary impulses in hypervectors with binary values.

%Bit-stream-based computing, also known as %techniques, %systems, 
Bit-stream computing, also known as stochastic computing (SC), %and unary computing (UC), 
has been the subject of a large body of recent %current
research efforts 
due to attractive advantages such as very-low implementation cost and high tolerance to noise~\cite{alaghi_survey}. %Considering bit-stream computing, 
SC operates on %processes
random sequences of binary bits, called \textit{bit-stream}. %sequences 
%instead of conventional binary radix data. %with weighted binary radix;
Similar to hypervectors in HDC, stochastic bit-streams are holographic with no bit significance. %weight. 
Complex arithmetic operations are simplified to basic logic operations in SC. %consisting of individual logic gates;
%thereby, reducing the arithmetic operations into the order of few logic gates. 
For instance, multiplication can be performed %is realized %can be performed 
using a single \texttt{AND} gate~\cite{10040569}. However, the accuracy of SC operations is severely affected by the random fluctuations in the bit-streams. Some operations, such as multiplication, similarity or correlation between bit-streams further degrades the quality of results. Often very long bit-streams need to be processed for accurate results. 
%The long length of bit-streams leads to long latency and high energy consumption. 
Recently, \textit{low-discrepancy (LD)} bit-streams were suggested to improve the quality of SC operations while reducing the length of bit-streams%by removing the random fluctuations error
~\cite{Alaghi_Fast, Sobol_TVLSI_2018, 7927069}. Logic-$1$s and logic-$0$s are uniformly spaced in these bit-streams. Hence, the streams do not suffer from random fluctuations. The correlation issue is further addressed with these bit-streams by using different LD distributions. LD bit-streams have been recently used to perform completely accurate computations with SC logic~\cite{NajafiLiljaSobol}. %Instead of true-random or pseudo-random bit-streams, the data are converted to LD bit-streams. 
%But in contrast to conventional SC systems that generate psuedo-bit-streams by using pseudo-random number generators, 
The LD bit-streams are generated by using \textit{quasi-random numbers} such as \textit{Sobol} sequences.  %With LD bit-streams, 
%The bit-streams quickly and monotonically converge to the target value, producing accurate results with very short bit-streams. 
%Among various LD bit-streams, \textit{Sobol} sequence-based bit-streams have proven to work 
%however, recent works proved how Sobol low-discrepancy (LD) sequences perform accurate arithmetic operations for bit-stream processing.

This work takes advantage of the recent progress in the bit-stream generation of SC systems to improve the data encoding of HDC systems. %Melting both computing paradigms in one pot is the subject of this study. 
To the best of our knowledge, we have pioneered the optimization of Sobol sequences using an algorithm to enhance orthogonality for their application in HDC systems within the literature.
%We propose to use Sobol sequences in HDC systems for the first time in the literature. 
We draw an analogy between SC and HDC, combining the data encodings of the two emerging computing paradigms. There have been some recent efforts to integrate SC and HDC~\cite{StocHD, LeveragingSHDC, 10194306, StochasticHDBarisAksanli, COSMO}. 
%The prior works for the collaboration of SC and HDC are present in \cite{StocHD, LeveragingSHDC, StochasticHDBarisAksanli, COSMO}. 
Unlike these previous studies that use pseudo-random sequences for random vector generation, this work provides randomness and guarantees independence between hypervectors %bit-streams 
by using \textit{optimized} LD Sobol sequences \cite{10137195}. %in the bit-streams prefer demonstrating uncorrelated bit-streams with LD sequences different than random vector generation using pseudo-random bit-streams. 
We first inspect the Sobol sequences obtained from the MATLAB tool. %in our proposed scheme. 
The hypervector representation in HDC is similar to the bit-stream representation in SC; hence, we  %can 
utilize stochastic cross-correlation ($SCC$)~\cite{Alaghi_SCC1}, a metric %which is 
used for determining the similarity of stochastic bit-streams, in evaluating %the similarity between 
hypervectors. We produce %obtain 
a matrix of %contains
hypervectors from %the 
Sobol sequences. The $SCC$ metric is then used to measure the correlation of each hypervector pair, yielding a \textit{distance matrix}. Any row-column combination of the distance matrix shows the absolute deviation from the $SCC$=$0$,  indicating independent hypervectors. %the uncorrelated bit-streams. 
We propose an algorithm to select %the uncorrelated 
the best independent Sobol-based hypervectors. %of 
%generated by Sobol sequences.
We utilize the selected hypervectors %Sobol elements 
for vector encoding in an HDC system case study. %are utilized. 
We %finally 
apply our proposed scheme to language and headline classification problems~\cite{RahimiRecentLanguage, RahimiGitHubBasedPaper, KLEYKO2016169}. 
We further evaluate the hardware efficiency of the new encoding module for HDC systems. In summary, the main contributions are as follows:
\begin{itemize}
\item For the first time, we utilize optimized LD Sobol sequences in data encoding of HDC and unveil their potential performance.
\item We propose an algorithm for selecting independent hypervectors by %the uncorrelated hypervectors; thereby, 
utilizing $SCC$ metric.
\item We find %list 
the top-performing Sobol sequences %indexes 
for generating independent %orthogonal %the uncorrelated 
hypervectors.
\item We compare the performance of Sobol-based hypervectors with two traditional approaches of encoding hypervectors using 1) linear-feedback shift registers (LFSRs) and 2) the MATLAB random function. 
\item Our experimental results show an accuracy improvement of up to 10.79\% for text classification.
\item Our new encoder module exhibits significant savings in energy consumption and area-delay product.
%During the tests, we also simulated the linear feedback shift register (LFSR) for vector generation with two different approaches.
\end{itemize}

The rest of the paper is organized as follows: Section~\ref{background} presents some basic concepts of HDC and SC. Section~\ref{proposed_methodology} describes the proposed methodology and presents the optimization of the best independent sequence selection. Experimental results are presented in Section~\ref{tests_and_results} for the language and newspaper headline classification problems. Finally, Section~\ref{conclusion} concludes the paper.

\section{Background}
\label{background}

\subsection{Hyperdimensional Computing (HDC)}

HDC is a brain-inspired computational model based on the observation that the human brain operates on high-dimensional representations of data. Reasoning in this robust model of computation is done by measuring the similarity of hypervectors~\cite{StocHD}.
%In HDC, 
Hypervectors are $D$-dimensional sequences with $+1$ and $-1$ values (corresponding to \textit{logic-1} and \textit{logic-0} in hardware, respectively). 
Prior works on HDC target near-orthogonal hypervectors with random distribution and approximately the same number of $+1$ and $-1$. So, the threshold value ($T$) is set to $0.5$~\cite{RahimiGitHubBasedPaper}. In HDC applications, using hypervectors with long lengths (in the range of 10,000 lengths or more) is common to reduce the similarity between the encoded vectors and improve the quality of results.

%There are prior studies for efficient encoding using better orthogonality; however, design space exploration is required for this optimization-oriented problem. For instance, Basaklar et al. employed a genetic algorithm-based approach for the optimized hypervector design. 

%\section{Background}
%This section presents a summary of the HDC as a vector-symbolic architecture.

%The basic computing primitive is a \textit{vector} having binary values ($+1$ or $-1$) in binary HDC. 
The basic operations in HDC are %listed as 
multiplication ($\oplus$: logical \texttt{XOR}), %in logic processing), 
addition ($\Sigma$: bitwise population count), %in logic processing), 
and permutation ($\Pi$: shifting). %in logic processing). 
These operations are invertible and have linear time complexity. HDC systems first encode data with a proper technique according to the %depending on the
%for
classification or cognitive tasks. %Prior works have used %The literature has 
Spatial, temporal, and histogram-based encoding techniques are used in the literature~\cite{8490896}.
Encoders are divided into (i)~record-based and (ii)~$n$-gram-based approaches \cite{ParhiSurvey, 9780410}. %The first assigns 
The record-based approaches assign level hypervectors ($\boldsymbol{L}$, e.g., pixel intensity values in image processing application) and position hypervectors ($\boldsymbol{P}$, e.g., randomly generated vectors for pixel positions). Feature positions on data are encoded via $\boldsymbol{P}$s that are orthogonal to each other. On the contrary, level hypervectors are expected to have correlations between neighbors. The final hypervector is denoted as $\boldsymbol{\mathcal{H}} = \Sigma_{i=1}^{N} (\boldsymbol{L}_i \oplus \boldsymbol{P}_i)$, where $N$ is the feature size. The second category %encoder
utilizes $n$-gram-based statistics like those in natural language processing systems. These encoders use %In these encoders, 
rotationally permuted hypervectors, which are orthogonal to each other. %, are used. 
The final hypervector is $\boldsymbol{\mathcal{H}} = \boldsymbol{L}_1 \oplus \pi\boldsymbol{L}_2 \oplus \pi^{N-1}\boldsymbol{L}_N$, where $\pi^{n}$ denotes the $n$-times rotationally permuted $\boldsymbol{L}$. All samples in the training dataset are evaluated for $\boldsymbol{\mathcal{H}}$, and each contributes to the corresponding class hypervector, which is the \textit{trained model} of the overall system. During the inference, the test data is encoded ($\boldsymbol{h}$), and the similarity check is performed between each test query and the class hypervector \cite{9786145}. %The $n$-gram-based approach is utilized in our proposed encoding scheme with Sobol sequences, and the language classification problem is tested.
In our encoding scheme with Sobol sequences, we utilize the $n$-gram-based approach and test the language processing problem~\cite{RahimiGitHubBasedPaper}.

\begin{table}[t]
\footnotesize
\centering
\caption{Comparison of SC and HDC %Systems
}
\label{sc_hdc_comparison_table}
\setlength{\tabcolsep}{2pt}
\scalebox{0.85}{
\begin{tabular}{|c|c|c|} 
\hline
 & \textbf{SC} & \textbf{HDC} \\ 
\hline
\begin{tabular}[c]{@{}c@{}}\textbf{Atomic}\\\textbf{Building Block}\end{tabular} & \makecell{Bit-stream~\cite{Alaghi_Survey_2018} \\ (size of $N$)} & \makecell{Hypervector~\cite{8714821} \\ (size of $D$)} \\ 
\hline
\begin{tabular}[c]{@{}c@{}}\textbf{Data}\\\textbf{Representation }\end{tabular} & \multicolumn{1}{l|}{\begin{tabular}[c]{@{}l@{}}Unipolar or Bipolar Bit-streams~\cite{Alaghi_Survey_2018} \\ \ \ \ \ \ \ Unary Bit-streams~\cite{SchoberMAC} \\ \ \ \ Low-Discrepancy %Quasi-random
Bit-streams~\cite{NajafiLiljaSobol} \end{tabular}} & \begin{tabular}[c]{@{}c@{}}Random Hypervectors~\cite{RahimiGitHubBasedPaper}\end{tabular} \\ 
\hline
\textbf{Metric} & Stochastic
  Cross-Correlation ~\cite{Alaghi_SCC1}& \begin{tabular}[c]{@{}c@{}}Cosine Similarity~\cite{RahimiGitHubBasedPaper, 9586253} \\Dot Product~\cite{COSMO} \\Hamming Distance~\cite{ParhiSurvey} \\Overlap
Coefficient~\cite{ParhiSurvey} \end{tabular} \\ 
\hline
\begin{tabular}[c]{@{}c@{}}\textbf{Target}\\\textbf{Representation}\end{tabular} & \begin{tabular}[c]{@{}c@{}}Uncorrelated Bit-streams~\cite{Alaghi_SCC1} \\Correlated Bit-streams~\cite{SchoberMAC}
%(or correlated if\\unary processing)
\end{tabular} & \makecell{Orthogonal Hypervectors~\cite{KLEYKO2016169}} \\
\hline
\end{tabular}
}
\end{table}

\subsection{%Bit-stream
Stochastic Computing (SC)}
%Bit-stream computing 
SC is a re-emerging paradigm that uses the power of processing random bit-streams %unconventional computing 
to reduce the complexity of arithmetic operations to the level of individual %a few standard 
logic gates~\cite{Alaghi_Survey_2018, Najafi_TVLSI_2019}. 
%Previous studies have broadly explained the strengths of this unconventional computing paradigm~\cite{Alaghi_Survey_2018, Najafi_TVLSI_2019}. 
%superiority of random or unary bit-streams in the literature~\cite{Alaghi_Survey_2018, SchoberMAC}. 
Let $X \in \mathbb{Z}_0^{+}$ be a scalar value to be represented with a stochastic bit-stream. A bit-stream $\boldsymbol{X}$ of size $N$ has $P_X = \frac{X}{N}$ probability for the occurrence of 1s. %\textit{logic-1}s.
Unlike conventional binary radix, stochastic bit-streams are free of bit significance. The ratio of the number of 1s to the length of bit-stream determines the bit-stream value. For instance, $\boldsymbol{X1} = 10101010$ represents $P_{X1} = \frac{4}{8}$ and $\boldsymbol{X2} = 10111101$ represents $P_{X2} =\frac{6}{8}$. %Bit-stream computing is free from the bit positions, and there is no bit significance like in conventional binary radix. 
%For instance, any bit-stream with 50\% of 1s (e.g., 1100, 101010, 00001111). 
%For instance, $X1 = 4$ can be represented in $\boldsymbol{X1} = 10101010$ for $P_{X1} = \frac{4}{8}$. 
%This %way of 
%approach of representation is called unipolar encoding (UPE) supporting positive scalars and zero: $X \in \mathbb{Z}^{+}_{0}$. Considering negative scalars, bipolar encoding (BPE) is utilized, but in this work we work with UPE. 
%As a second scalar, let $X2 = 6$ and $\boldsymbol{X2}$ represent bit-stream $10111101$. 
Applying bit-wise \texttt{AND} %logic 
operation to these bit-streams produces %, $\boldsymbol{X1} \ \& \ \boldsymbol{X2}$, 
an output bit-stream $\boldsymbol{Y} = 10101000$ with %, is obtained having 
probability $P_Y = \frac{3}{8}$ that is equal to $P_{X1} \times P_{X2}$. %Likewise, \texttt{XNOR} gate works as a multiplier of BPE-based bit-streams. 
For correct functionality and accurate result, the two operand bit-streams need to be \textit{independent} or \textit{uncorrelated}.
%Both multiplications require uncorrelated bit-streams for an accurate result; 
Accurate conversion of scalar values to stochastic bit-streams while guaranteeing independence between them has been a long-time challenge in SC~\cite{Alaghi_SCC1}.  
%however, this may not always be the case due to the correlated bit-streams or random fluctuations in SC \cite{Alaghi_SCC1, AlaghiDissertation}. 
The state-of-the-art work has addressed this challenge by encoding data to \textit{LD bit-streams}~\cite{Najafi_TVLSI_2019, Sobol_TVLSI_2018}. The %scalar 
input data is compared with quasi-random numbers such as Sobol numbers from a Sobol sequence (please see~\ref{Append}\blue{ppendix}). The comparison output generates an LD bit-stream representing the input value. A 1 is generated if $scalar > random \ number$. A 0 is produced otherwise.  
%In the literature, prior state-of-the-art efforts have already proposed a way for solving inaccuracy problem by utilizing the LD quasi-random sequences~\cite{7927069, Sobol_TVLSI_2018, NajafiLiljaSobol}. 
%In traditional random bit-stream generation, a scalar to be encoded is compared to a random number, and the related bit is set if $scalar > random \ number$; otherwise, the related bit is \textit{logic-0}. 
For an $N$-bit long bit-stream, the input %scalar 
data is compared with $N$ Sobol numbers. %from the Sobol sequence. 
%a new random number is generated at each attempt of $N$ bits. In Sobol LD case, $N$ sequence elements are utilized for the comparison yielding quasi-randomness~\cite{NajafiLiljaSobol}.
\autoref{sc_hdc_comparison_table} compares the SC and HDC computational models.

%In conclusion, stochastic bit-stream processing and HDC systems are compared in \autoref{sc_hdc_comparison_table}.  

\section{Proposed Methodology}
\label{proposed_methodology}

\subsection{From Bit-Streams to Hypervectors}
\begin{figure}[t]
  \centering
  \includegraphics[width=\linewidth]{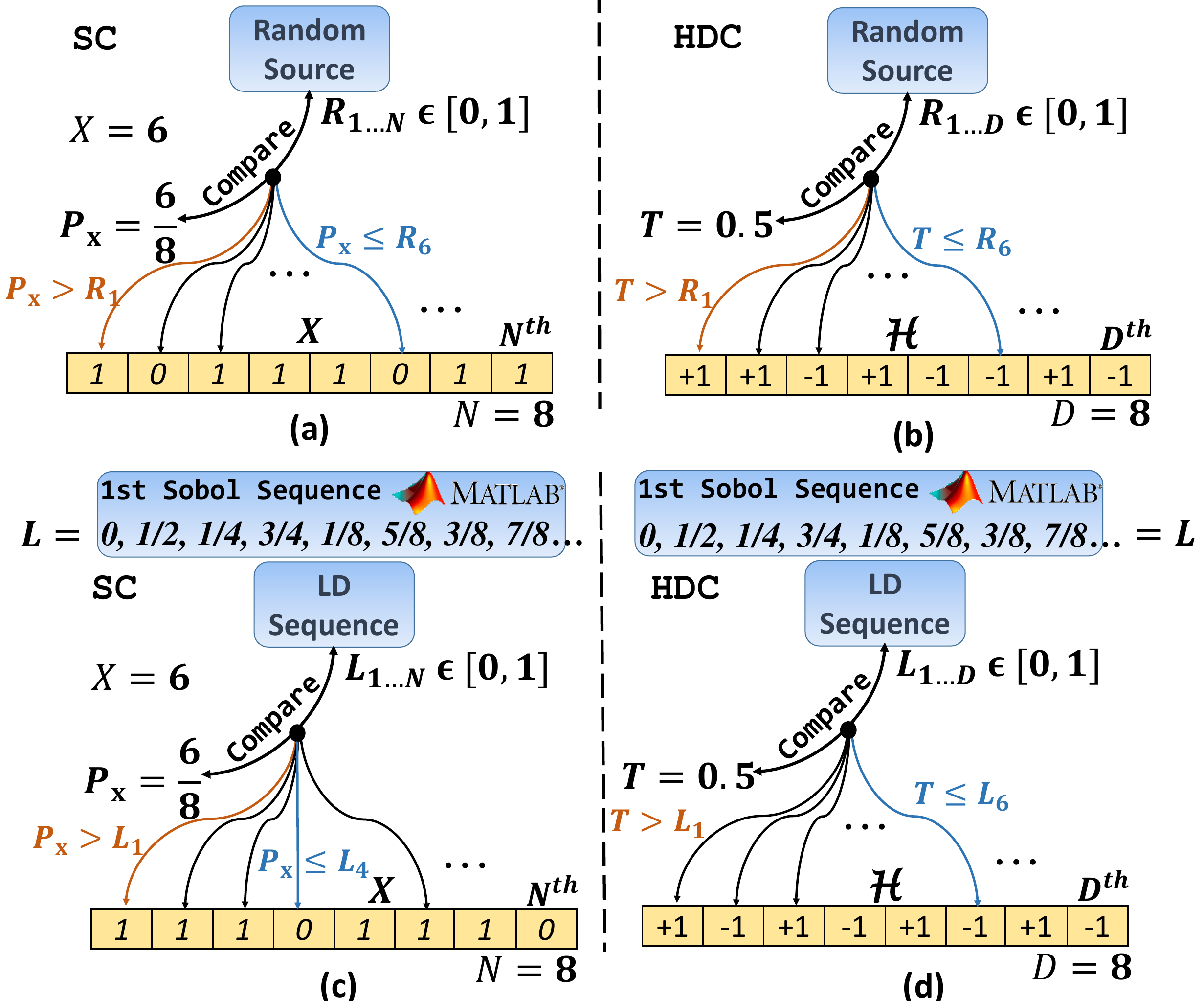}
  \caption{(a) Conventional random source-based stochastic bit-stream generation, (b) traditional hypervector generation using a random source, (c) LD bit-stream generation using a Sobol sequence, and (d) utilizing Sobol sequences for generating hypervectors.   %The Sobol LD sequence usage for SC-based systems is exemplified in (c), while Sobol utilization for HDC systems is proposed in (d).
  }
  \label{sc_versus_hdc}
\end{figure}

SC and HDC %systems have similar 
both exploit a redundant holographic data representation. While conventional binary radix assigns weight %significance 
to each bit depending on its significance, %for scalar representation, 
SC and HDC systems utilize unweighted %long 
sequences of binary bits~\cite{AygunGunesCT, 8714821}. %holding binary values free from the position. 
%In HDC, hypervectors are $D$-dimensional sequences with $+1$ and $-1$ values (corresponding to \textit{logic-1} and \textit{logic-0}, respectively). 
%In SC, bit-streams are $N$-bit binary sequences of 1s and 0s encoding a probability value. 
In both computational model, the encoding includes %there is 
a comparison with a random %reference
value, $R$. A \textit{scalar} value in SC and a \textit{threshold} value in HDC are the actors of this comparison.
Figs.~\ref{sc_versus_hdc}(a) and (b) show examples of the traditional approaches for encoding data in SC and HDC.
%The examples of data encoding in SC and HDC are presented in Fig.~\ref{sc_versus_hdc}, together with the Sobol LD utilization proposal in Fig.~\ref{sc_versus_hdc} (d). 
In Fig.~\ref{sc_versus_hdc}(a), the scalar value $X$ is encoded to a bit-stream of size $N=8$  representing %according to 
the probability $P_X$. The random source in Figs.~\ref{sc_versus_hdc}(a) and (b) is a number generator that generates  random numbers %reference value
in the $[0,1]$ interval. 
%for each bit. 
Random vector generation for HDC is shown in Fig.~\ref{sc_versus_hdc}(b). 
Here, the threshold value ($T$) is
$0.5$. 
%In prior work, the threshold value, $T$, is set to $0.5$, and $+1$ and $-1$ values are randomly distributed~\cite{RahimiGitHubBasedPaper}. 
%randomly distributed in half are targeted \cite{RahimiGitHubBasedPaper}.
Fig.~\ref{sc_versus_hdc}(c) shows how data is encoded to an LD bit-stream by exploiting a \textit{Sobol} sequence.

\ignore{
 %shows the use of Sobol LD sequences for data representation in SC. 
The MATLAB built-in Sobol sequence generator~\cite{MATLABsobolset} can be used to efficiently generate Sobol arrays. The procedure for generating Sobol numbers is as follows: 
%The production of Sobol arrays obtained with MATLAB~\cite{MATLABsobolset} is as follows:
{\itshape
\noindent According to Joe and Kuo~%'s remarks in 
\cite{SobolMath}, any $j^{th}$ component of the points in a Sobol sequence is generated by first defining a primitive polynomial $x^{s_j} + a_{1,j}x^{{s_j}-1}+...+a_{s_j-1,j}x + 1$ of a degree of $s_j$ in the field $\mathbb{Z}_2$. Any `$a$' satisfies $a \in \{0,1\}$. By considering bit-by-bit \textit{XOR} operator, $\oplus$, the `$a$' coefficients are utilized for a sequence $\{m_{1,j}, m_{2,j},...\}$ by a relation given as $m_{k,j} = 2a_{1,j}m_{{k-1},j} \oplus 2^2a_{2,j}m_{{k-2},j} \oplus ... \oplus 2^{{s_j}-1}a_{{s_j}-1,j}m_{k-{s_j}+1,j} \oplus 2^{{s_j}}m_{k-{s_j},j} \oplus m_{k-{s_j},j}$. The $m$ values can be arbitrarily chosen provided that $1 \leq k \leq s_j$, and $m_{j,k} \in \{2n+1: n \in \mathbb{Z}_0^{+} \} $ and $m_{j,k} < 2^k$. With a denote of direction numbers, $\{v_{1,j}, v_{2,j}, ... \}$, where any $v_{j,k} = \frac{m_{k,j}}{2^k}$, $j^{th}$ component of the $i^{th}$ point in a Sobol sequence is presented: $x_{i,j} = b_1v_{1,j} \oplus b_2v_{2,j} \oplus...$, where any b is the right-most bits (i.e., least-significant ones) of the $i$ sub-index in binary form.
}
}

The produced $N$-dimensional Sobol arrays {with recurrence relation} are used as an ideal random source to generate \textit{accurate} LD bit-streams. %for bit-stream generation, just like a random source, as shown in Fig.~\ref{sc_versus_hdc} (c).
Sobol sequences have also been successful in providing the needed \textit{independence} between stochastic bit-streams. Generating different LD bit-streams by using different Sobol sequences is sufficient to guarantee %the needed 
independence between bit-streams~\cite{NajafiLiljaSobol}. 
%quasi-randomness is successful in nearly-uncorrelated bit-stream generation and provides more accurate arithmetic results in SC. 
%At this point, we propose the use of HDC hypervectors, as shown in Fig.~\ref{sc_versus_hdc} (d).
Motivated by the success of using quasi-random numbers in SC, this work employs Sobol sequences for generating %near-ideal 
hypervectors of HDC systems.
Fig.~\ref{sc_versus_hdc}(d) presents the idea. We use %LD 
Sobol sequences as the random source to generate \textit{uncorrelated} %\textit{orthogonal} 
hypervectors with \textit{desired ratio} of $+1$ and $-1$. 
%Like the random source in conventional systems shown in Fig.~\ref{sc_versus_hdc} (b), the LD sequence is a source of reference values but targets the production of ``uncorrelated" hypervectors. 
Conventionally, HDC systems sets the \textit{threshold} value ($T$) to 0.5 to generate hypervectors with 50\% $+1$ and 50\% $-1$.  
Unlike prior HDC systems, this work explores a range of values for $T$ to achieve the best accuracy with Sobol-based hypervectors.
%Unlike conventional HDC systems, we do not use only the $T$=$0.5$; and an iteration is applied for the best accuracy of a system with Sobol-based hypervectors. 
The challenging optimization problem is to determine the best set of %independent 
Sobol sequences for any $T$ value. 
%Thus, for any $T$,  the determination of the best-uncorrelated sequences is a challenging optimization problem. 
The %monitoring 
merit metric for this optimization is $SCC$ as given in equation~(\ref{SCC_equation}):
\begin{equation}
\label{SCC_equation}
    SCC = \begin{cases} \frac{ad-bc}{D \times min(a+b, a+c)-(a+b) \times(a+c)} & ,  \  if \ ad>bc \\ \frac{ad-bc}{(a+b) \times(a+c) - D \times max(a-d, 0)} & ,  \ else\end{cases}
\end{equation}

\begin{figure}[t]
  \centering
  \includegraphics[width=\linewidth]{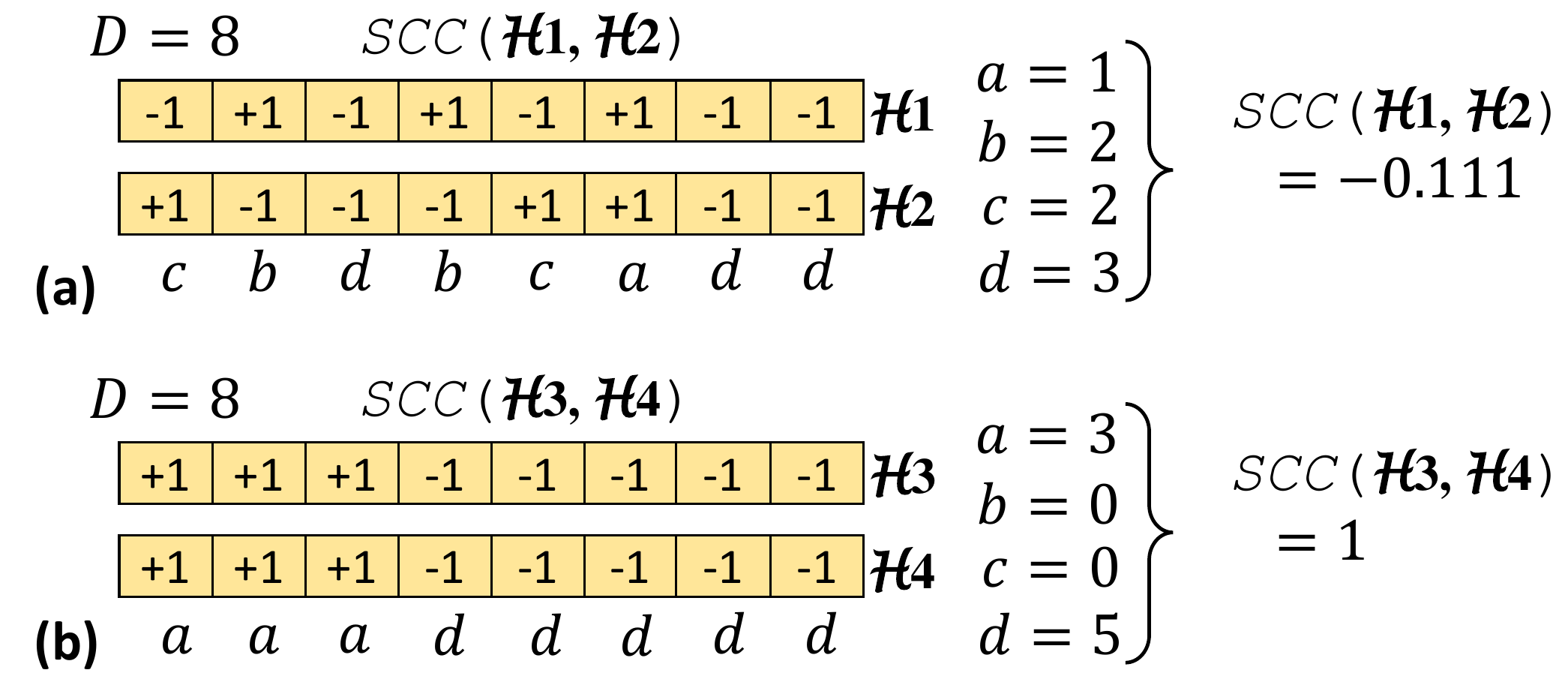}
  \caption{Correlation measurement of two sample hypervectors: (a) near-zero correlation, (b) highly correlated. %example.
  }
  \label{scc_example} 
\end{figure}

The $a$, $b$, $c$, and $d$ variables in %denoted in 
the $SCC$ equation \cite{Alaghi_SCC1} %metric 
are the cumulative sum of overlaps between two hypervectors: 
{$a = | \{ \boldsymbol{\mathcal{H}x}_i = \boldsymbol{\mathcal{H}y}_i = +1 \}|$, $b = | \{ \boldsymbol{\mathcal{H}x}_i = +1, \boldsymbol{\mathcal{H}y}_i = -1 \}|$, $c = | \{ \boldsymbol{\mathcal{H}x}_i = -1, \boldsymbol{\mathcal{H}y}_i = +1 \}|$, $d = | \{ \boldsymbol{\mathcal{H}x}_i = \boldsymbol{\mathcal{H}y}_i = -1 \}|$.}
%$a$=$(+1, +1)$, $b$=$(+1, -1)$, $c$=$(-1, +1)$, and $d$=$(-1, -1)$. %values from the two hypervectors. 
%The $SCC$ metric returns 
$SCC$ is a value in the $[-1,+1]$ interval. %between $+1$ and $-1$. 
A zero or near-zero $SCC$ %value indicates 
means \textit{uncorrelated} %or orthogonal 
hypervectors. %$SCC$=$0$ is the perfect uncorrelated vector measurement, and 
$SCC$=$+1$ indicates a positive correlation (totally similar), while $SCC$=$-1$ shows a negative correlation (no overlap). 
Fig.~\ref{scc_example} exemplifies two pairs of hypervectors and their corresponding $SCC$ values.
%The use of $SCC$ metric for any two hypervectors is shown in Fig.~\ref{scc_example}. 
All hypervectors here have a probability $3/8$ of observing $+1$.
%Two hypervectors, with $3/8$ probability of getting $+1$s, are considered for correlation measurement. 
The $SCC$ value is calculated by finding the $a$, $b$, $c$, and $d$ values.
%After getting $a$, $b$, $c$, and $d$ values, $SCC$ formula returns the correlation coefficient. 
The example in Fig.~\ref{scc_example}(a) includes two hypervectors with near-zero $SCC$, while the hypervectors in Fig.~\ref{scc_example}(b) are identical and so have $SCC$=$1$.

{Table~\ref{new_SCC_COS} presents the $SCC$ and cosine (cos) similarity metrics for Cartesian pairs of hypervectors %obtained by 
generated using %through
MATLAB's built-in Sobol generator. For $T$=$0.5$, we observed a high number of orthogonal pairs. %even though the complete orthogonality percentage is observed high, 
The $SCC$ and cos metric intervals were both $[0,1]$ for $T$=$0.5$. This indicates that some vector pairs are also completely correlated ($SCC$=$cos$=$1$). For $T$=$0.3$, the number of  orthogonal pairs was low.  %Although the complete orthogonal vector occurrence is low for $T$=$0.3$, 
However, the $SCC$ and cos interval was narrower, i.e., far from the correlated case ($SCC$=$1$), swinging around zero. $SCC$ fluctuates in $\sim[-0.2, 0.4]$ for this threshold value. %, also giving information about negative correlation.
We will show that the best results are achieved when $T$ is close to $0.3$.
%$T$ selection considering both metrics is crucial for Sobol-based encoding, and $T$=$0.5$ as a frequently utilized threshold in the literature may create a fallacy in the Sobol case due to the presence of its fully-correlated vectors.
}

%In Fig.~\ref{scc_example} (a) a near-zero example is given, and in Fig.~\ref{scc_example} (b), the totally identical hypervectors, $\boldsymbol{\mathcal{H}3}$ and $\boldsymbol{\mathcal{H}4}$, are exemplified for $SCC$=$1$.

%Thus, unlike the prior art, not directly the orthogonality of vectors (with cosine similarity and dot product metrics); but the uncorrelation is measured with $SCC$. 

\begin{table}[t]
\caption{$SCC$ and Cos Similarity-Based Threshold Monitoring*}
\label{new_SCC_COS}
\scalebox{0.99}{
\begin{tabular}{|c|l|c|c|c|} 
\hline
\multicolumn{2}{|c|}{\textbf{$\mathbf{SCC}$ and Cos Sim. }} & \textbf{$D$=2048} & \textbf{$D$=4096} & \textbf{$D$=8192} \\ 
\hline
\multirow{2}{*}{$T$=0.5} & $SCC$ Range & {[}0,1] & {[}0,1] & {[}0,1] \\ 
\cline{2-5}
 & Cos Range & {[}0,1] & {[}0,1] & {[}0,1] \\ 
\hline
\multicolumn{1}{|l|}{\multirow{2}{*}{$T$=0.3}} & $SCC$ Range & {[}-0.19, 0.43] & {[}-0.17, 0.43] & {[}-0.16, 0.43] \\ 
\cline{2-5}
\multicolumn{1}{|l|}{} & Cos Range & {[}0, 0.52] & {[}0, 0.52] & {[}0, 0.52] \\ 
\hline
\multicolumn{5}{l}{\footnotesize{*Tests over the first 100 Sobol sequences in MATLAB}}
\end{tabular}
}
\end{table}

\setlength{\textfloatsep}{0.35cm}
\setlength{\floatsep}{0.35cm}
\begin{algorithm}[t]
\caption{Sobol-based $\boldsymbol{\mathcal{H}}ypervectors$ Generation}
\label{Algorithm_1}
\begin{algorithmic}[1]
\algsetup{linenosize=\small}
\small
\REQUIRE $Sobol_{1111 \times D}$, $T: threshold$, \\ $D: \boldsymbol{\mathcal{H}}ypervectors \ size$,
\ENSURE $\boldsymbol{\mathcal{H}}ypervectors$
\FOR{$i=1:1:1111$}
\FOR{$j=1:1:D$}
\IF {$T \leq Sobol_{1111}(i,j)$}
\STATE $\boldsymbol{\mathcal{H}}ypervectors(i,j) = -1$
\ELSE
\STATE $\boldsymbol{\mathcal{H}}ypervectors(i,j) = +1$
\ENDIF
\ENDFOR
\ENDFOR
\RETURN $\boldsymbol{\mathcal{H}}ypervectors$
\end{algorithmic}
\end{algorithm}

\setlength{\textfloatsep}{0.35cm}
\setlength{\floatsep}{0.35cm}
\begin{algorithm}[t]
\caption{$SCC$ of $\boldsymbol{\mathcal{H}}ypervectors$ Cartesian Product}
\label{Algorithm_2}
\begin{algorithmic}[1]
  \algsetup{linenosize=\small}
  \small
\REQUIRE $\boldsymbol{\mathcal{H}}ypervectors$,
\ENSURE $VAL$, $IDX$, $Distance$
\FOR{$i=1:1:1111$}
\FOR{$j=1:1:1111$}
\STATE $Distance(i,j) \leftarrow$  \STATE $|$\texttt{SCC}$\big(\boldsymbol{\mathcal{H}}ypervectors(i,:), \ \boldsymbol{\mathcal{H}}ypervectors(j,:)\big)|$
\ENDFOR
\ENDFOR
\STATE $[VAL, \ IDX] \leftarrow$ \texttt{sortMin}$\big($\texttt{sumColumns}($Distance$)$\big)$
\RETURN $VAL$, $IDX$, $Distance$
\end{algorithmic}
\end{algorithm}

\subsection{On the Decision of the Best-Performing Sobol Sequences}
An important optimization problem in the HDC literature is to select the best orthogonal hypervectors.  
%The selection of the best uncorrelated vectors based on Sobol sequences is considered as an optimization problem in this work. 
Algorithm~\ref{Algorithm_1} shows the procedure for generating Sobol-based hypervectors. 
%In Algorithm~\ref{Algorithm_1}, Sobol-based hypervector generation is presented. 
We use the MATLAB tool %platform 
and its built-in Sobol sequence generator~\cite{MATLABsobolset}, which implements %based on 
%above-mentioned remarks of
Joe and Kuo's method~\cite{SobolMath} as discussed in the \ref{Append}\blue{ppendix}. The maximum number of Sobol sequences that MATLAB can produce is 1111. %$1111$.
In Algorithm~\ref{Algorithm_1}, $D$ is the hypervector size. Thereby, the $Sobol$ matrix has a size of $1111 \times D$. 
All $Sobol$ numbers are compared with the $T$ value, and the results ($-1$ or $+1$) are recorded in a $\boldsymbol{\mathcal{H}}ypervectors$ matrix of $1111 \times D$ size.
%Based on the $T$ value, all elements of $Sobol$ are compared and the result is recorded in $\boldsymbol{\mathcal{H}}ypervectors$ matrix in $1111 \times D$ size.

\begin{figure*}
  \centering
  \includegraphics[width=525pt, height=250pt]{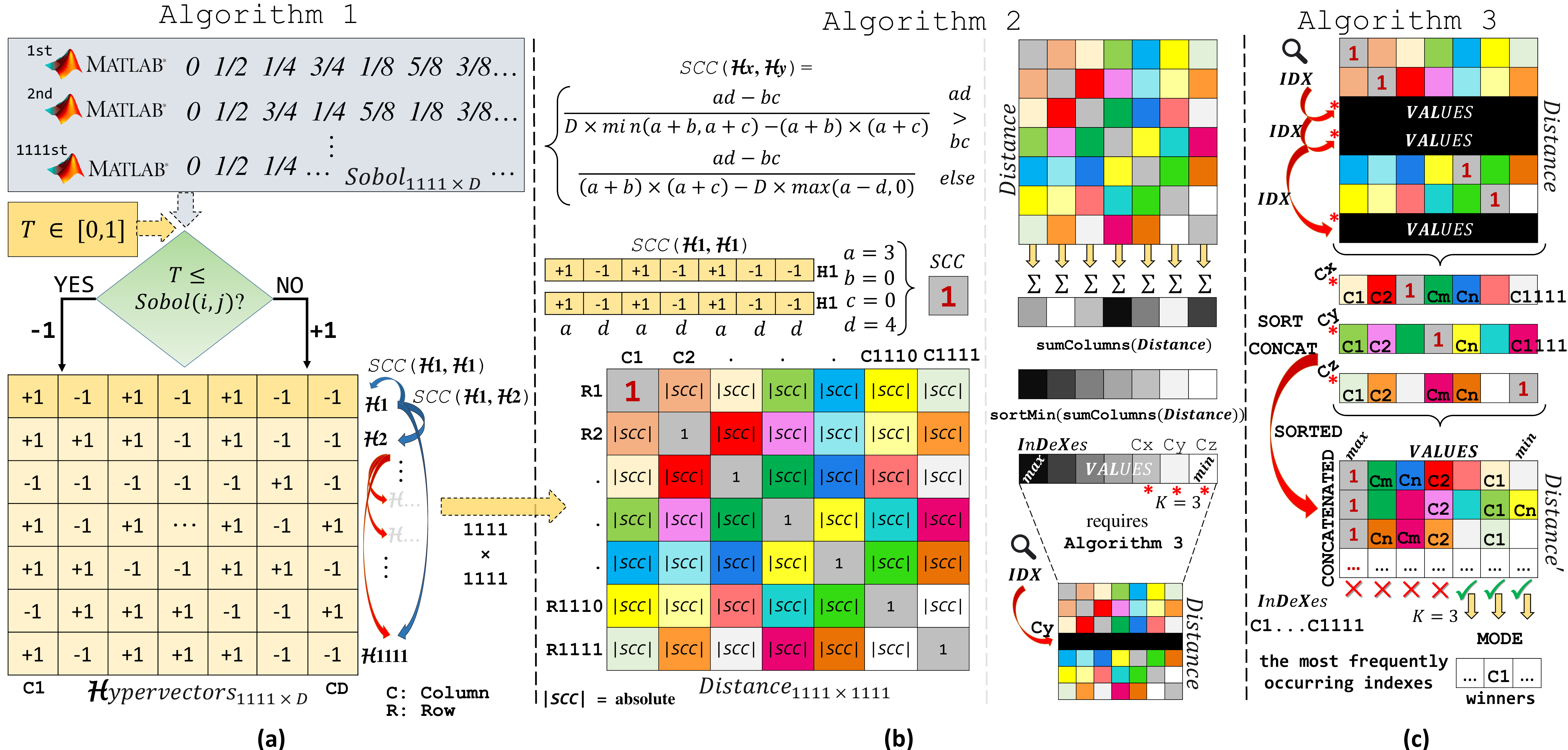}
  \caption{The illustration of (a) Algorithm~\ref{Algorithm_1}, (b) Algorithm~\ref{Algorithm_2}, and (c) Algorithm~\ref{Algorithm_3}. %\red{[Sercan, can you bold the font of "sumColumns(Distance)" and other small texts in the figures]} \blue{I have updated Dr Najafi}. Thanks Sercan.
  }
  \label{all_agorithms}
  \vspace{+5pt}
\end{figure*}

Algorithm~\ref{Algorithm_1} returns the $\boldsymbol{\mathcal{H}}ypervectors$ matrix. Algorithm~\ref{Algorithm_2} %requires 
uses this matrix to generate %for the generation of
an $SCC$-based $Distance$ matrix of Cartesian products.
%: $\boldsymbol{\mathcal{H}} \times \boldsymbol{\mathcal{H}} = \{(\boldsymbol{\mathcal{H}x}, \boldsymbol{\mathcal{H}y}) | \boldsymbol{\mathcal{H}x} \in \boldsymbol{\mathcal{H}}, \boldsymbol{\mathcal{H}y} \in \boldsymbol{\mathcal{H}} \}$.
Each pair of $\boldsymbol{\mathcal{H}x}$ and $\boldsymbol{\mathcal{H}y}$ in $\boldsymbol{\mathcal{H}}ypervectors$
%Each two couples in $\boldsymbol{\mathcal{H}}ypervectors$, $\boldsymbol{\mathcal{H}x}$ and $\boldsymbol{\mathcal{H}y}$, 
is compared using the $SCC$ metric, yielding the $Distance$ matrix size of $1111 \times 1111$ that holds the absolute values of $SCC$. 

\setlength{\textfloatsep}{0.45cm}
\setlength{\floatsep}{0.45cm}
\begin{algorithm}[t]
\caption{Minimum of Minima}
\label{Algorithm_3}
\begin{algorithmic}[1]
  \algsetup{linenosize=\small}
  \small
\REQUIRE $IDX$, $Distance$, $K$
\ENSURE $SobolUncorrelated$
\FOR{$i=1:1:K$}
\STATE $minOFmin \leftarrow$ \texttt{sortMin}$\big(Distance(IDX(i),:)\big)$
%\STATE $minOFmin \leftarrow minOFmin(1:K)$ 
\STATE $Concatenated_{K \times D} \leftarrow$ \texttt{CONCAT}($minOFmin$)
\ENDFOR
\STATE $SobolUncorrelated \leftarrow$ \texttt{MODE}($Concatenated_{K \times D}.IDX$, $K$)
\RETURN $SobolUncorrelated$
\end{algorithmic}
\end{algorithm}

%\vspace{+5pt}

\begin{figure*}[t]
%\vspace{-1em}
  \centering
  \includegraphics[width=515pt, height=250pt]{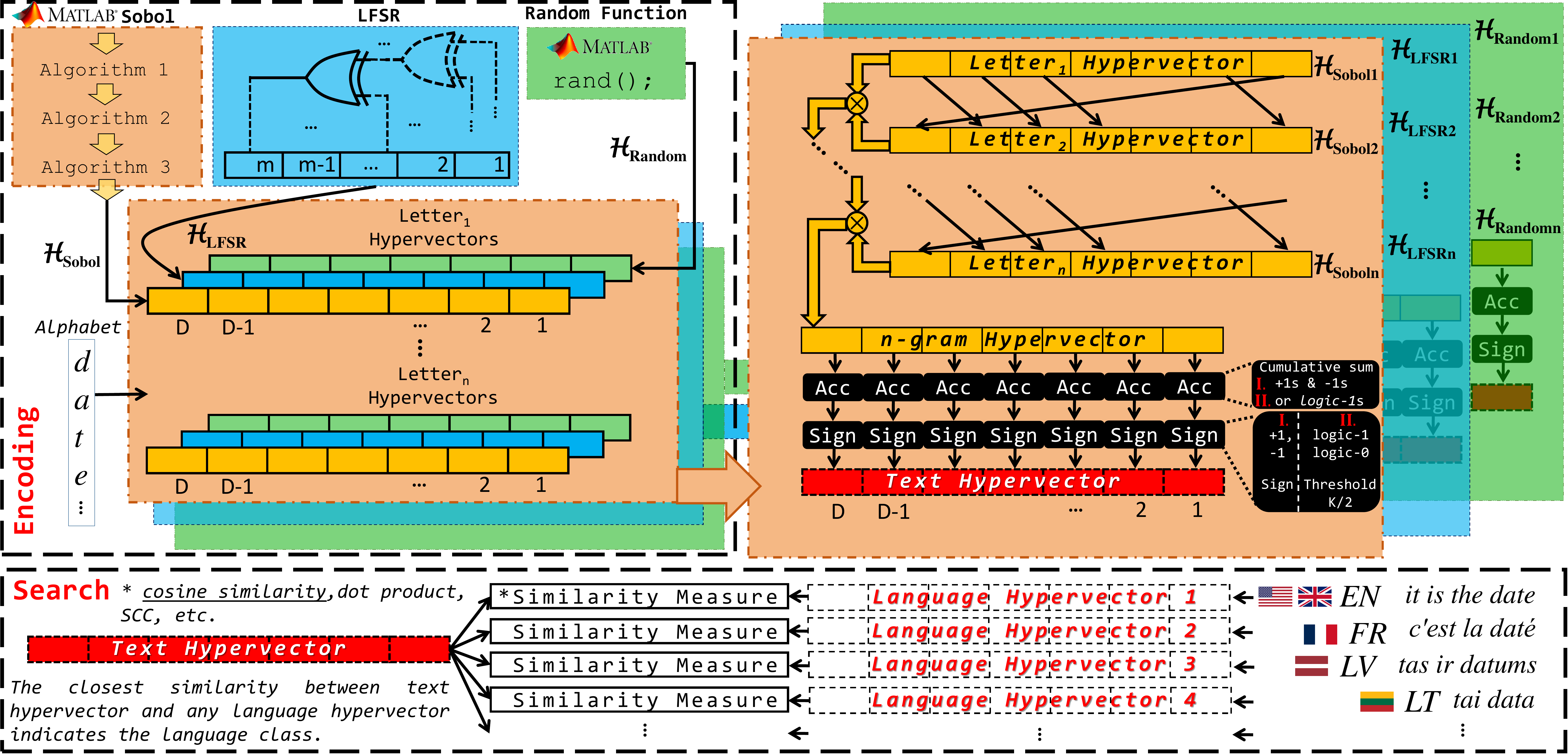}
  %\vspace{-0.5em}
  \caption{The overall architecture of the HDC language classifier.}
  \label{proposed_architecture}
    %\vspace{-10pt}
\end{figure*}

%In Algorithm~\ref{Algorithm_2}, 
Lines $1$ to $6$ of Algorithm~\ref{Algorithm_2} build %set 
the $Distance$ matrix by calculating the $SCC$ values. %show the related $SCC$ calculation and setting $Distance$ matrix.
Any $i^{th}$ row and $j^{th}$ column intersection holds the correlation coefficient between the $i^{th}$ and $j^{th}$ hypervectors in the $\boldsymbol{\mathcal{H}}ypervectors$ matrix. The $Distance$ matrix is a symmetric, square matrix holding $SCC = 1$s in the diagonal elements (%, since any 
$SCC(\boldsymbol{\mathcal{H}x}, \boldsymbol{\mathcal{H}x})=1$.) %returns 1. 
After producing the $Distance$ matrix, the algorithm %Algorithm~\ref{Algorithm_2} 
performs column-wise operations to select the minimum distances at each column ({similarly, row-wise operations are possible} %or row is also possible 
due to symmetry). Line 7 of Algorithm~\ref{Algorithm_2} first calculates the summation of each Distance matrix column, and then sorts the results of summations. %shows that \red{each column is summed.} 
%\hassan{I don't get this.}
%\sercan{Each column is the Sobol ID, or each row, too, due to the symmetry. Algorithm 2 line 7 make the adding any column values vertically reducing to a vector. This supplies the each Sobol ID's total distance performance w.r.t. others. And then we make a sorting, again like in the Line 7 of Algo. 2. Maybe re-writing like:\red{Line 7 of Algorithm 2 calculates vertical summation of each Distance matrix columns, then sorting the obtained sum. }}
%\hassan{Let me put it this way. Why "each" and "columns"? This is what I don't get. Do you mean each "column"?}
%\sercan{Yes Dr. Najafi, it is each column, I am sorry for using column as plural. This is the exact part of the Algorithm what I want to indicate with this sentence: https://www.hizliresim.com/scyadrb (a link for a figure) }
The summation is a vector with $1 \times D$ elements. Minimum sorting is applied to find the values and indexes of Sobol elements with minimum $SCC$.  Figs.~\ref{all_agorithms}(a) and (b) depict Algorithms~\ref{Algorithm_1} and~\ref{Algorithm_2},  respectively. As it can be seen, after obtaining the $\boldsymbol{\mathcal{H}}ypervectors$ matrix by comparing Sobol numbers with $T$, %with a $T$ comparison, 
$SCC$ measures %calculates 
the correlation of the Cartesian-based hypervector pairs. % couples. 
The $Distance$ matrix is symmetric, and each color represents the absolute value of an $SCC$. 
%Correlation of any hypervector with itself gives $1$ ($SCC(\boldsymbol{\mathcal{H}1}, \boldsymbol{\mathcal{H}1})$=1). These set the diagonal.  
Fig.~\ref{all_agorithms}(b) shows how column-wise summation is performed in Algorithm~\ref{Algorithm_2},  yielding symbolic gray-tone representation. After sorting, the \textit{values} ($VAL$s) and \textit{indexes} ($IDX$s) are recorded. The top-$K$ minimum values %($VAL$s) 
of these elements are called the minima of the $Distance$ matrix, and the corresponding $IDX$s %will be 
are used %utilized 
for the \textit{minimum of minima} in Algorithm~\ref{Algorithm_3}, as depicted in Fig.~\ref{all_agorithms}(c).

Algorithm~\ref{Algorithm_3} %finds and 
returns a $SobolUncorrelated$ vector, which holds the $K$ best uncorrelated 
Sobol %element 
indexes. The inputs of the algorithm are
%In Algorithm~\ref{Algorithm_3}, 
$IDXs$ and the $Distance$ matrix from Algorithm~\ref{Algorithm_2}, %are required 
%together with 
%in addition to
beside $K$, the number of to-be-selected Sobol indexes. %The return value for the algorithm is $SobolUncorrelated$ vector holding the best uncorrelated Sobol element indexes. 
The top-$K$ %selected
minimum distances and their indexes ($IDX$s) from Algorithm~\ref{Algorithm_2} are %considered 
used here. %considered.
In Fig.~\ref{all_agorithms}(c), the selected $IDX$s of the $Distance$ matrix are pointed with black-colored rows. Each row is further processed for the minimum $SCC$s. %in the $Distance$. 
The $for \ loop$ in  Algorithm~\ref{Algorithm_3} processes the $K$ rows by \texttt{sort}ing and \texttt{CONCAT}aneting them, %yielding 
producing a new $Distance'$ matrix with a size of $K \times D$. Finally, the $Distance'$ matrix is checked with the \texttt{MODE} function to return %, which returns 
the most repetitive $IDX$s of the top-$K$ minimum-valued %$Distance'$ 
columns. Column-wise frequent index check allows us to see how repetitive Sobol elements are %present 
in the minima of $SCC$ distances. Fig.~\ref{all_agorithms}(c) depicts Algorithm~\ref{Algorithm_3} for $K$=3; the top-$K$=3 minimum of minima are selected for the %final winners in
$SobolUncorrelated$ vector. If any repetitive $IDX$ occurs, $K$ is increased for the next available Sobol element %winner 
to keep the column list unique in $SobolUncorrelated_{1 \times K}$.

%In Algorithm~\ref{Algorithm_3}, after sorting the $Distance$ matrix top-$K$ selected rows based on $IDX$s, the sorted rows are concatenated in line 4. The frequent index selection, i.e., minimum of minima, is presented in line 6 of Algorithm~\ref{Algorithm_3}. The \texttt{MODE} procedure works on the $IDX$s of $Distance'$ matrix. Again, top-$K$ minimum of minima are selected for final the winners in $SobolUncorrelated$. Fig.~\ref{all_agorithms}(c) depicts Algorithm~\ref{Algorithm_3} for $K=3$. 

\vspace{+5pt}

\subsection{The Overall Architecture}
In this section, we present the overall HDC architecture. % before the tests.
Without loss of generality, we apply the proposed technique to a word-processing HDC system for language classification. In this system, a hypervector is needed for each alphabet letter. %element.
%We first target the word processing for language classification; therefore, we require the hypervectors for each alphabet element. 
Our approach uses Sobol-based hypervectors. Nonetheless, we also implement the system with the MATLAB random function and an LFSR-based random number generator for comparison purposes.
%Our proposal is to use Sobol-based hypervector; nonetheless, we also work on the random function- and LFSR-based number generators for comparison. 
The overall architecture is shown in Fig.~\ref{proposed_architecture}. For each letter in the alphabet, hypervector generation is performed with our Sobol-based technique, LFSR, and the MATLAB built-in random function \texttt{rand()}, %(built-in MATLAB \texttt{rand()} function), 
%yielding 
producing $\boldsymbol{\mathcal{H}_{Sobol}}$, $\boldsymbol{\mathcal{H}_{LFSR}}$, and $\boldsymbol{\mathcal{H}_{Random}}$, respectively. The $n$-gram  approach \cite{ParhiSurvey}, is applied to the incoming data, i.e., $n$ consecutive letters. %Sobol-based generation is provided with proposed algorithms. 
For the LFSR-based approach, random numbers are generated  %considering
using the maximal-period LFSRs described in~\cite{PhilipKoopmanLFSR} %polynomials in~\cite{PhilipKoopmanLFSR} 
for each length $D$. 
The initial seed value of the LFSRs is randomly selected. %assigned 
%during the generation of hypervectors.
%Considering 
Similar to Rahimi et al.'s HDC architecture in~\cite{RahimiGitHubBasedPaper} and \cite{Rahimigithub}, the calculation of an $n$-gram hypervector is done by rotating the letter hypervectors, %is 
as shown in Fig.~\ref{proposed_architecture}. We keep a copy of the hypervectors generated by the three approaches 
%\hassan{What do you mean by "a copy of the three approaches''? do you mean a copy of the hypervectors generated by these approaches?}
%\sercan{Yes Dr. Najafi, this is "a copy of the hypervectors generated by these approaches" }
%(Sobol, LFSR, and random) 
for this process. Then, the $n$-gram hypervector is accumulated.
%PREVIOUS VERSION BEFORE RE_WRITTEN
%\red{The accumulation operation (\texttt{Acc}) can be an algebraic operation for $+1$, $-1$, or a logic process for the population count of \textit{logic-1}s in actual hardware. After that, the thresholding compared to the $K/2$ threshold for the latter is applied. For the former,} 
%\hassan{Sercan, I have a hard time understating these. Can you re-write these few red sentences?}
%\sercan{Re-written version:}

During text hypervector generation, the \textit{accumulation} operation (\texttt{Acc}) is an algebraic summation of $+1$s and $-1$s %values 
in hypervectors. %for $+1$, $-1$ valued hypervectors, and 
The \textit{thresholding} is %then 
applied via the \texttt{sign} function. In %actual 
hardware, the population count of \textit{logic-1}s is used %applied 
for %the 
\textit{accumulation}, %which is 
followed by a \textit{thresholding} operation by comparison with $K/2$. %, %via $K/2$ comparison 
%\hassan{Do you mean comparison with K/2?}
%\sercan{Yes, Dr. Najafi. Half size of the K total hypervector amount is the reference to decide +1, -1 values in the text hypervector.}
%as presented in Fig.~\ref{proposed_architecture}. 
We proceed with the former approach in our simulations by using algebraic \textit{accumulation} and  \texttt{sign}-based \textit{thresholding}. After iterating over the incoming data to generate the text hypervectors, %are iterated  for the text hypervector generation, %finally, 
the classification is performed in the search module by performing %utilizing the 
a similarity check between the text hypervector and the language hypervector.

\begin{table*}
\centering
\caption{Hardware efficiency considering the embedded platform, CPU, GPU, and ASIC design}
\scalebox{1}{
\begin{tblr}{
  cells = {c},
  cell{1}{1} = {c=2}{},
  cell{1}{3} = {c=2}{},
  cell{1}{5} = {c=2}{},
  cell{1}{7} = {c=2}{},
  cell{1}{9} = {c=2}{},
  cell{3}{1} = {r=2}{},
  vlines,
  hline{1-3,5} = {-}{},
  hline{4} = {2-10}{},
  stretch = 0
}
\textbf{\textit{Performance }} &  & {\textbf{(i) Performance }\\\textbf{in an Embedded}\\\textbf{Platform (ARM) }} &  & {\textbf{(ii) CPU Workload }\\\textbf{Performance}\\\textbf{(Intel i5)}} &  & {\textbf{(iii) GPU Power Load~}\\\textbf{Performance~}\\\textbf{(NVIDIA~Quadro 6000) }} &  & {\textbf{(iv) Encoding Module }\\\textbf{ASIC Design} \\\textbf{(45 $nm$)}} & \\
D & \textbf{\textit{Encoding }} & \textbf{\textit{Runtime }} & \textbf{\textit{Memory }} & \textbf{\textit{Average}} & \textbf{\textit{Max.}} & \textbf{\textbf{\textit{Average}}} & \textbf{\textbf{\textit{Max.}}} & \textbf{\textit{Energy }} & \textbf{\textit{Area$\times$Delay}}\\
$8192$ & Random & 1,068.3sec & 18.3KB & 10.6\%(idle+9\%) & 15.2\%(idle+13.2\%) & $0.354mW$ & $0.402mW$ & $16.88 nJ$ & $721.12$$\times$$10^{-9}$\\
 & Sobol & 687.4sec & 17.8KB & 9.4\%(idle+7.4\%) & 12.1\%(idle+10.9\%) & $0.036mW$ & $0.256mW$ & $2.69pJ$ & $59.20$$\times$$10^{-12}$
\end{tblr}
\label{new_hardware}
}
%\vspace{-0.5em}
\end{table*}

\section{Tests and Results}
\label{tests_and_results}

%\hassan{Sercan, something that came to my mind is, how about adding a table to compare the performance of our technique with LFSR and Random method in generating "Orthogonal" hypervectors. In Table I, we discuss 4 metrics for HDC. I don't know if we have enough time to prepare these numbers but a reviewer may wonder why we don't compare the performance for the needed "orthognality". Now we don't present any result to support that the hypervectors generated with our technique are completely orthognal. We have good results for the application/case study part. But before going to application we may still show the quality of hypervectors with HDC metrics. Dot product for example, is a good metric to show the orthognality. }
%\sercan{Table II has been added to mitigate the issue on this}

\subsection{Hardware Performance}
{ We first evaluate the hardware efficiency of the proposed encoder module (all designs are in $D$=$8192$). We use four hardware workspaces: (i) ARM-based embedded platform (a resource-limited device with 700 MHz, 32-bit, single-core), (ii) Central processing unit (CPU - Intel(R) Core(TM) i5-10600K $@$4.10GHz), (iii) Graphics processing unit (GPU - NVIDIA Quadro RTX 6000), and (iv) and an ASIC design with 45 $nm$ technology.
The first workspace considers two complete HDC systems with a random function-based approach and a Sobol-based approach. The overall system was implemented in C language and deployed into the ARM processor. The random method dynamically creates data with a built-in C language-based \texttt{rand} function. On the other hand, the Sobol sequences are pre-generated, stored in, and read from memory. Table~\ref{new_hardware} presents the performance (i.e., run-time and memory usage) results. The presented results are based on the single-time hypervector assignments for each letter; however, the training phase requires iteration for the random method that severely worsens during training. Table~\ref{new_hardware} also shows the CPU-based workload performances of the encoding part of an HDC system. We iteratively ($10^7$ times) create an alphabet with letter hypervectors and compare the workload brought by the random-based and Sobol-based approaches in the CPU. The iterations guarantee fairness in terms of possible background tasks and processes. We also put an idle time by waiting for stabilization before initiating each run. Table~\ref{new_hardware} shows the average and maximum of all iterations. CPU workload during idle time and total increments on the hypervector operations (\textit{idle} + \textit{workload by vector generation} $\%$) are recorded. The Sobol-based approach exhibits less load compared to the random case. GPU power consumption for the hypervector operations is also considered. The Sobol-based approach provides nearly $10$ times lower power consumption. Last but not least, we implement an ASIC hypervector encoding module to evaluate the energy and area-delay product. For a complete alphabet generation, especially targeting the %further 
architectures for training-on-edge, an encoding module is designed. The random approach uses LFSRs, and the Sobol approach reads Sobol sequences from block random access memory. The Sobol approach performance in both energy consumption and the area-delay product brings substantial outcomes and is promising for next-generation computing systems like HDC.
}

\begin{table}[t]
\centering
%\renewcommand{\arraystretch}{1.15}
%\caption{The Overall Evaluation Results}
\caption{Classification Rates for Different Encoding Methods}
\label{full_results}
\scalebox{0.94}{
\begin{tabular}{|p{0.15cm}|p{2.7cm}|p{1cm}|p{1cm}|p{1cm}|p{0.2cm}|} 
\hline
\vcell{\textbf{\textit{D}}} & \vcell{\begin{tabular}[b]{@{}c@{}}\textbf{Encoding}\\\textbf{Methods}\end{tabular}} & \vcell{\begin{tabular}[b]{@{}c@{}}\textbf{Min.}\\\textbf{Acc.}\end{tabular}} & \vcell{\begin{tabular}[b]{@{}c@{}}\textbf{Max.}\\\textbf{Acc.}\end{tabular}} & \vcell{\begin{tabular}[b]{@{}c@{}}\textbf{Std.}\\\textbf{Dev.}\end{tabular}} & \vcell{\begin{tabular}[b]{@{}c@{}}\textbf{Avg.}\\\textbf{Acc.}\end{tabular}} \\[-\rowheight]
\printcellbottom & \printcellmiddle & \printcellmiddle & \printcellmiddle & \printcellmiddle & \printcellmiddle \\ 
\hline
\parbox[t]{2mm}{\multirow{5}{*}{\rotatebox[origin=c]{90}{16}}}
 & Random Vector & 13.19\% & 18.14\% & 0.0108 & 15.07\% \\ 
\cline{2-6}
 & LFSR w/ Random Seed & 12.45\% & 18.20\% & 0.0119 & 15.17\% \\ 
\cline{2-6}
 & \multicolumn{1}{c}{Sobol LD Sequence} & \multicolumn{3}{c}{-} & \textbf{17.80\%} \\
 & \multicolumn{5}{l|}{Sobol $IDX$s ($T$=0.66): 2, 3, 8, 11, 13, 14, 16, 18, 27, 28, 29, 34,} \\
 & \multicolumn{5}{l|}{35, 40, 42, 44, 46, 47, 50, 52, 53, 54, 55, 60, 61, 62, 63, 66} \\ 
\hline
\parbox[t]{2mm}{\multirow{5}{*}{\rotatebox[origin=c]{90}{32}}}
& Random Vector & 20.47\% & 25.70\% & 0.0121 & 22.71\% \\ 
\cline{2-6}
 & LFSR w/ Random Seed & 18.69\% & 25.70\% & 0.0121 & 22.47\% \\ 
\cline{2-6}
 & \multicolumn{1}{c}{Sobol LD Sequence} & \multicolumn{3}{c}{-} & \textbf{26.52\%} \\
 & \multicolumn{5}{l|}{Sobol $IDX$s ($T$=0.30): 2, 4, 7, 8, 12, 20, 25, 27, 30, 29, 36, 40,} \\
 & \multicolumn{5}{l|}{42, 39, 43, 58, 51, 52, 64, 66, 74, 46, 79, 69, 72, 80, 65, 88} \\ 
\hline
\parbox[t]{2mm}{\multirow{5}{*}{\rotatebox[origin=c]{90}{64}}}
 & Random Vector & 31.07\% & 39.52\% & 0.0151 & 35.43\% \\ 
\cline{2-6}
 & LFSR w/ Random Seed & 31.78\% & 37.10\% & 0.0124 & 34.57\% \\ 
\cline{2-6}
 & \multicolumn{1}{c}{Sobol LD Sequence} & \multicolumn{3}{c}{-} & \textbf{46.22\%} \\
 & \multicolumn{5}{l|}{Sobol $IDX$s ($T$=0.70): 3, 6, 14, 16, 18, 19, 25, 26, 28, 30, 31, 35,} \\
 & \multicolumn{5}{l|}{36, 41, 54, 49, 73, 61, 67, 64, 72, 90, 91, 86, 88, 96, 100, 101} \\ 
\hline
\parbox[t]{2mm}{\multirow{5}{*}{\rotatebox[origin=c]{90}{128}}}
& Random Vector & 48.63\% & 54.80\% & 0.0127 & 52.04\% \\ 
\cline{2-6}
 & LFSR w/ Random Seed & 48.02\% & 54.02\% & 0.0112 & 51.16\% \\ 
\cline{2-6}
 & \multicolumn{1}{c}{Sobol LD Sequence} & \multicolumn{3}{c}{-} & \textbf{60.74\%} \\
 & \multicolumn{5}{l|}{Sobol $IDX$s ($T$=0.74): 2, 4, 5, 7, 14, 11, 13, 16, 23, 24, 26,} \\
 & \multicolumn{5}{l|}{25, 32, 34, 36, 42, 39, 53, 50, 48, 61, 67, 70, 64, 66, 72, 78, 84} \\ 
\hline
\parbox[t]{2mm}{\multirow{5}{*}{\rotatebox[origin=c]{90}{256}}}
& Random Vector & 67.36\% & 71.70\% & 0.0103 & 69.24\% \\ 
\cline{2-6}
 & LFSR w/ Random Seed & 66.80\% & 71.67\% & 0.0105 & 68.61\% \\ 
\cline{2-6}
 & \multicolumn{1}{c}{Sobol LD Sequence} & \multicolumn{3}{c}{-} & \textbf{79.03\%} \\
 & \multicolumn{5}{l|}{Sobol $IDX$s ($T$=0.70): 2, 3, 16, 10, 25, 28, 19, 38, 32, 60, 45,} \\
 & \multicolumn{5}{l|}{49, 39, 53, 54, 68, 80, 82, 64, 90, 78, 95, 99, 112, 120, 108, 92, 118} \\ 
\hline
\parbox[t]{2mm}{\multirow{5}{*}{\rotatebox[origin=c]{90}{512}}}
& Random Vector & 82.32\% & 83.83\% & 0.0039 & 83.03\% \\ 
\cline{2-6}
 & LFSR w/ Random Seed & 81.31\% & 83.18\% & 0.0045 & 82.22\% \\ 
\cline{2-6}
 & \multicolumn{1}{c}{Sobol LD Sequence} & \multicolumn{3}{c}{-} & \textbf{89.47\%} \\
 & \multicolumn{5}{l|}{Sobol $IDX$s ($T$=0.70): 2, 8, 9, 13, 16, 19, 21, 28, 29, 48, 32,} \\
 & \multicolumn{5}{l|}{44, 38, 64, 50, 51, 54, 58, 61, 85, 73, 74, 97, 101, 112, 90, 117, 120} \\ 
\hline
\parbox[t]{2mm}{\multirow{5}{*}{\rotatebox[origin=c]{90}{1024}}}
& Random Vector & 90.27\% & 91.51\% & 0.0025 & 91.15\% \\ 
\cline{2-6}
 & LFSR w/ Random Seed & 90.01\% & 91.22\% & 0.0028 & 90.56\% \\ 
\cline{2-6}
 & \multicolumn{1}{c}{Sobol LD Sequence} & \multicolumn{3}{c}{-} & \textbf{93.78\%} \\
 & \multicolumn{5}{l|}{Sobol $IDX$s ($T$=0.70): 1, 4, 9, 14, 26, 22, 21, 31, 51, 30, 34, 36,} \\
 & \multicolumn{5}{l|}{61, 71, 67, 73, 97, 99, 100, 96, 107, 125, 108, 110, 152, 120, 150, 140} \\ 
\hline
\parbox[t]{2mm}{\multirow{5}{*}{\rotatebox[origin=c]{90}{2048}}}
& Random Vector & 94.73\% & 95.40\% & 0.0015 & 95.14\% \\ 
\cline{2-6}
 & LFSR w/ Random Seed & 94.21\% & 95.04\% & 0.0017 & 94.71\% \\ 
\cline{2-6}
 & \multicolumn{1}{c}{Sobol LD Sequence} & \multicolumn{3}{c}{-} & \textbf{96.31\%} \\
 & \multicolumn{5}{l|}{Sobol $IDX$s ($T$=0.34): 6, 5, 9, 14, 22, 16, 47, 51, 55, 64, 58, 68, 78,} \\
 & \multicolumn{5}{l|}{87, 105, 88, 97, 96, 100, 166, 129, 132, 145, 152, 153, 114, 179, 164} \\ 
\hline
\parbox[t]{2mm}{\multirow{5}{*}{\rotatebox[origin=c]{90}{4096}}}
& Random Vector & 96.68\% & 97.09\% & 9.20e-04 & 96.88\% \\ 
\cline{2-6}
 & LFSR w/ Random Seed & 96.44\% & 96.88\% & 0.0012 & 96.68\% \\ 
\cline{2-6}
 & \multicolumn{1}{c}{Sobol LD Sequence} & \multicolumn{3}{c}{-} & \textbf{97.05\%} \\
 & \multicolumn{5}{l|}{Sobol $IDX$s ($T$=0.34): 1, 4, 6, 16, 30, 21, 26, 48, 33, 55, 43, 78,} \\
 & \multicolumn{5}{l|}{60, 62, 82, 96, 97, 87, 88, 91, 92, 126, 100, 105, 109, 115, 121, 127} \\ 
\hline
\parbox[t]{2mm}{\multirow{5}{*}{\rotatebox[origin=c]{90}{8192}}}
& Random Vector & 97.47\% & 97.87\% & 8.78e-04 & 97.68\% \\ 
\cline{2-6}
 & LFSR w/ Random Seed & 97.36\% & 97.73\% & 9.19e-04 & 97.55\% \\ 
\cline{2-6}
 & Multiple LFSRs & \multicolumn{3}{c|}{-} & 97.31\% \\ 
\cline{2-6}
 & \multicolumn{1}{c}{Sobol LD Sequence} & \multicolumn{3}{c}{-} & \textbf{97.85\%} \\
 & \multicolumn{5}{l|}{Sobol $IDX$s ($T$=0.38): 2, 5, 12, 15, 23, 36, 48, 51, 53, 54, 63, 73, 66,} \\
 & \multicolumn{5}{l|}{79, 97, 115, 88, 98, 104, 109, 159, 148, 147, 123, 126, 130, 188, 172} \\
\hline
\end{tabular}
}
\end{table}

\begin{figure*}[t]
  \centering
  \includegraphics[width=485pt, height=210pt]{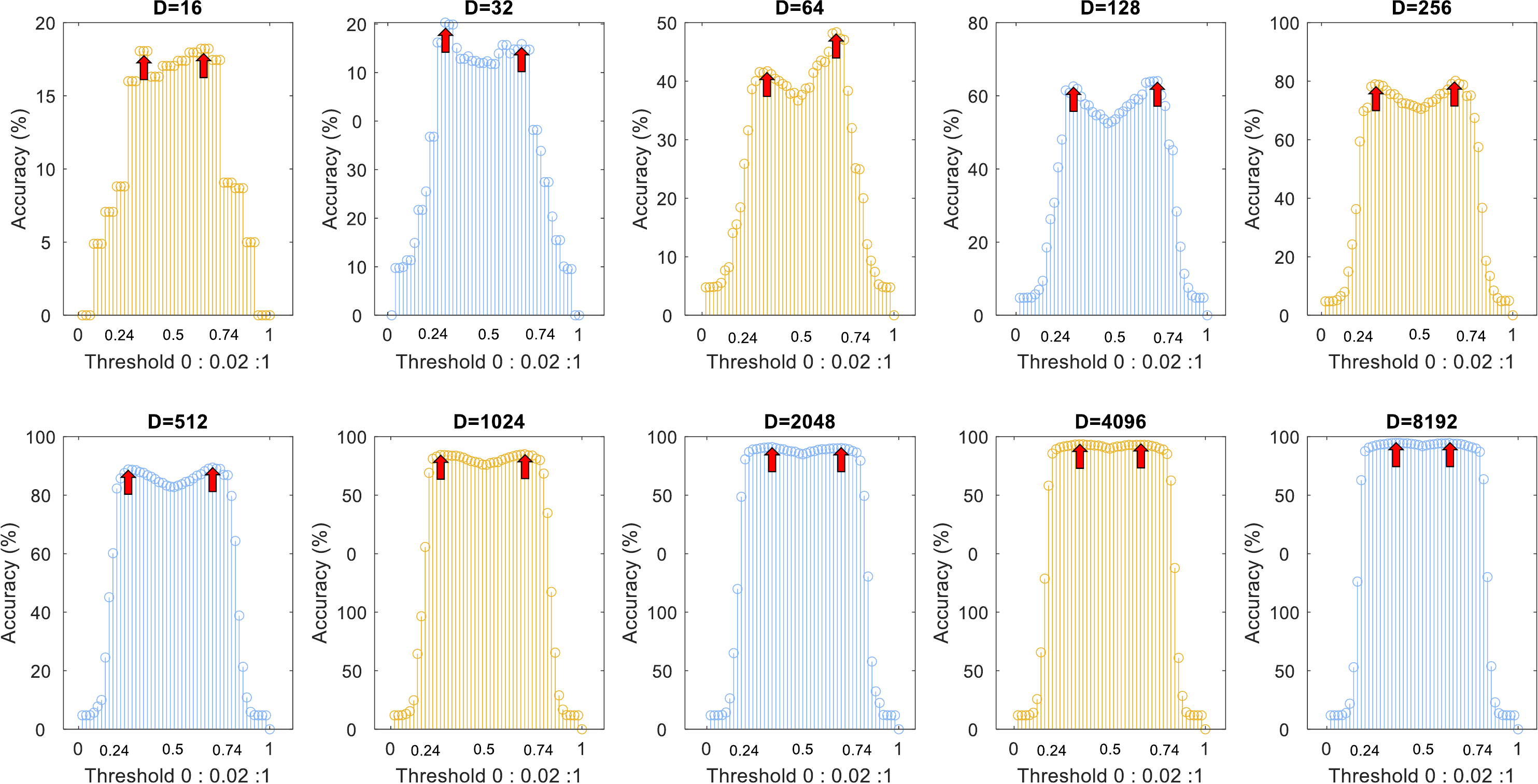}
  \caption{{Preliminary analysis of $T$ for different hypervector sizes ($D$) with the first 28 Sobol sequences of MATLAB (\includegraphics[width=5pt]{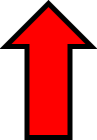}: maximum accuracy point).}
  %The decision of the $T$.   Preliminary analysis via validation accuracy with Top-$K$=$28$ first Sobol sequences.
  }
  \label{t_inspection}
\end{figure*}

\begin{figure}[t]
  \centering
  \includegraphics[width=\linewidth]{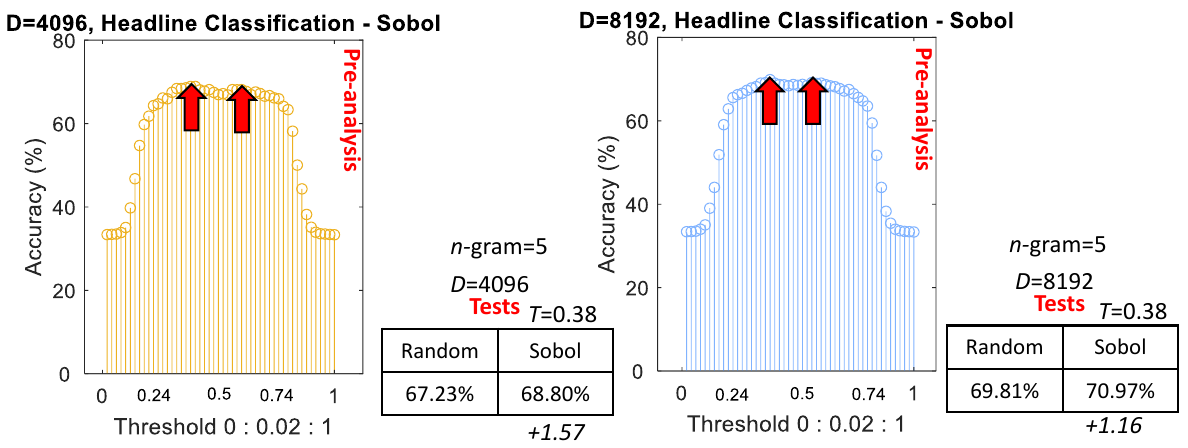}
  \caption{Preliminary analysis of $T$ and tests for the headline dataset.% and tests for $D$=$4096$ and $D$=$8192$.
  }
  \label{t_inspection_test_headline}
\end{figure}

\subsection{Accuracy}

For accuracy evaluation, we used two different datasets for separate performance monitoring: the 21-class European languages dataset~\cite{Quasthoff2006CorpusPF}
%for training 
and the newspaper headlines~\cite{headlinedataset} dataset. Following the testing approach of~\cite{RahimiGitHubBasedPaper} and \cite{ Rahimigithub}, %same testing style in \cite{RahimiGitHubBasedPaper, Rahimigithub}, 
first, the Europarl Parallel Corpus dataset~\cite{Europarl} was used %utilized 
for the inference %test 
step. The selected $n$-gram was four %$4$ 
for better accuracy, as reported in \cite{RahimiGitHubBasedPaper}. First, the training dataset was pre-analyzed over a $1000$-element validation set with the MATLAB tool's first $28$ %built-in 
Sobol sequences. Hence, %thereby, 
$K$=$28$ ($26$ letters, one space, and one extra character). Fig.~\ref{t_inspection} presents the pre-analysis for $T$. %, considering the validation accuracy. 
The $T$ values on the \texttt{x}-axis range from $0$ to $1$ with $0.02$ steps. %($T \leftarrow 0 : 0.02 : 1$), where 
The hypervector size $D$ %is 
varies from $16$ to $8192$. The preliminary analysis for each $D$ size %value 
helps us determine the approximate $T$ range that gives the %highest 
maximum accuracy %for 
with the Sobol sequences. We observed that, whether the first $K$ Sobol sequences are used, or the $K$ Sobol sequences are randomly selected, %or random $K$ indexes of the Sobol sequence are selected, 
the relative analysis of the $T$ values shows similar distributions, and the peak accuracy is obtained around $T$=$0.34$ ($\pm0.04$) or $T$=$0.7$ ($\pm0.04$). %After that,
The selected $T$s are %then
%utilized 
used in Algorithm~\ref{Algorithm_1} %for 
with the best uncorrelated Sobol sequences obtained by Algorithms~\ref{Algorithm_2} and~\ref{Algorithm_3}. 

\newpage

%After $T$ determination for each $D$,
After determining the best $T$ for each $D$, tests are performed %for each $D$ 
with different encoding methods (Sobol, LFSR, and random). Table~\ref{full_results} presents the results. 
%The results are presented in Table~\ref{full_results}. 
For the \textit{LFSR}-based encoding, we evaluated all maximal-period LFSRs corresponding to each $D$~\cite{PhilipKoopmanLFSR}. For the \textit{random} encoding, we run 1000 trials, each %trial
generating a different set of random numbers. We report the minimum, maximum, and average accuracy for the \textit{LFSR} and \textit{random} encoding.  %Sobol, LFSR, and random approaches are reported for each $D$ value. 
%The built-in MATLAB function provides Sobol sequences. LFSR is obtained by the model of maximal-length polynomials~\cite{PhilipKoopmanLFSR} in software model, and random is provided by the MATLAB built-in \texttt{rand()} function (recalling Fig.~\ref{sc_versus_hdc}). 
As it can be seen, the Sobol-based encoding %results 
achieves %show 
superior performance in all cases. %tests. 
The case with $D$=$8192$ delivers the best accuracy %better accuracy
($97.85\%$), close to the baseline accuracy from a conventional machine learning approach%reported in
~\cite{RahimiGitHubBasedPaper}. %by reducing the gap between the baseline accuracy ($99.8\%$), which has the conventional machine learning approach specified in \cite{RahimiGitHubBasedPaper}. 
%When looking at the results, 
The second-best outcomes are obtained with the hypervectors produced with the MATLAB \textit{random} function, %encoding 
and then the hardware-friendly LFSR-based encoding provides the lowest accuracy. %software-based random function (average of $1000$ trials). while the LFSR, an actual hardware model, gives the lowest results. 
%For LFSR, the initial seed value is randomly assigned during the generation of each letter hypervector, and a fixed maximal-length LFSR polynomial is used for the corresponding $D$ value.
%Additionally, for $D$=$8192$, $K$=$28$ different LFSR models are also produced independently; however, we report that the LFSR result is lower when the seed value is kept constant. 
Table~\ref{full_results} reports the best Sobol $IDX$s selected with % the help of 
the proposed algorithms. %presented. 
%During the sorting procedure with the proposed algorithms, the priority is given to the lower indexes among $IDX$s with the same $SCC$ distance. 
%; this is the reason for reporting low $IDX$s in Table~\ref{full_results}.

%\newpage

To show %prove 
the superiority of the proposed encoding with another dataset, %our proposal in another dataset, 
we also tested the classification problem of newspaper headlines on three topics (\textit{entertainment}, \textit{politics}, and \textit{parenting}) obtained from the HuffPost newspaper headlines released in Kaggle~\cite{headlinedataset}. This is a relatively more complex problem with shorter headlines compared to paragraphs. For training, %the training set, 
$3400$ headlines were utilized for each class, while $1000$ different headlines were used for inference. %testing. 
Fig.~\ref{t_inspection_test_headline} shows the pre-analysis of $T$ for this %the headline 
dataset for $D$=$4096$ and $8192$. %the similar $T$ inspection like the previous tests with the validation pre-analyses. 
As can be seen, a similar distribution is obtained by finding %getting 
the best accuracy %higher validation accuracy at 
around $T$=$0.34$ ($\pm0.04$).
For $n$-gram=$5$ and $D$=$8192$, the \textit{random} function-based approach showed an accuracy of $69.81\%$ %presented a $69.81\%$ test accuracy 
(average of $1000$ trials), while the proposed \textit{Sobol}-based method %proposal 
achieved an accuracy of $70.97\%$. %for $D$=$8192$. 

These findings underscore the significance of the optimized Sobol-based architecture. The design is enhanced through optimization algorithms that leverage meticulously chosen, best orthogonal Sobol random sources. It should be emphasized that HDC systems generally adhere to a single-pass learning strategy, presenting results based on scanning the dataset only once without a backward pass. On the other hand, to establish edge-compatible training, the raw dataset has been presented without the use of any additional feature extraction or % and without requiring 
multiple iterations based on learning rate or error optimization. %Particularly, complex 
Conventional neural network systems with complex %many %a crowded structure requiring
matrix multiplications pose bottlenecks for edge devices during error optimization with many %necessary 
partial derivative calculations, learning rate-based fine-tuning, and batch processing in multi-stochastic data processing.
{
{
In contrast, HDC systems offer a %much more 
lightweight solution for the same accuracy level. %iso-accuracy. %For instance, 
A neural network-based system established for similar accuracy is more costly and less efficient in hardware. The proposed %optimized 
vector generation technique %introduces 
opens a new perspective for hardware-friendly learning. %systems. 
It is lightweight, highly accurate, and requires %In this regard, an extremely lightweight learning methodology has been pursued, providing a high-accuracy option with optimized Sobol and an encoding strategy requiring 
only one round of hypervector generation. %In literature,
In prior HDC systems,
selecting the best vectors involved multiple iterations with a random source to achieve good orthogonality \cite{10248004}. In contrast, LD Sobol sequences %do not require this and 
inherently provide the needed orthogonality; with our optimization approach, higher quality is guaranteed %optimized version further guarantees higher-quality 
for hypervector generation.
}
}

{
{
\section{Conclusion}
\label{conclusion}
In this work, we introduced a novel, lightweight %machine learning 
approach featuring an innovative encoding technique for generating high-quality hypervectors %tailored 
for hyperdimensional computing (HDC) systems. Inspired by recent strides in low-discrepancy encoding methods proposed for stochastic computing (SC) systems, we employed quasi-random Sobol sequences, coupled with an optimization framework, to produce orthogonal %uncorrelated 
hypervectors with varied distributions and ratios of +1s and -1s. Our methodology exploits %followed a meticulously designed 
an optimization algorithm to identify the optimal set of Sobol sequences, minimizing correlations crucial for vector symbolic data processing. To substantiate the effectiveness of the proposed technique, we conducted a comprehensive performance evaluation, comparing the method with two conventional approaches for hypervector generation based on LFSRs and algorithmic random functions. We evaluated the proposed approach for %Our proposal encompasses diverse encoding strategies within 
a letter processing HDC system, scrutinizing accuracy and integrating hardware designs across four distinct processing environments: ARM embedded device, CPU,  GPU, and custom ASIC design. Our novel encoding technique demonstrated superior classification accuracy across varied datasets. It also showed higher %by emphasizing the imperative of edge-compatible system designs for hardware platforms.
%Furthermore, it showcased enhanced 
hardware efficiency, considering factors such as energy efficiency and area-delay product.
}
}

\section*{\sc Acknowledgments}
This work was supported in part by National Science Foundation (NSF) grant \#2019511, the Louisiana Board of Regents Support Fund \#LEQSF(2020-23)-RD-A-26, and generous gifts from Cisco, Xilinx, and NVIDIA.

%\section{Appendix}
\appendix[Formal Definition of Sobol Sequences]
\label{Append}
The MATLAB built-in Sobol sequence generator~\cite{MATLABsobolset} can be used to efficiently generate Sobol arrays. The procedure for generating Sobol numbers is as follows: 
%The production of Sobol arrays obtained with MATLAB~\cite{MATLABsobolset} is as follows:
{\itshape
\noindent According to Joe and Kuo~%'s remarks in 
\cite{SobolMath}, any $j^{th}$ component of the points in a Sobol sequence is generated by first defining a primitive polynomial $x^{s_j} + a_{1,j}x^{{s_j}-1}+...+a_{s_j-1,j}x + 1$ of a degree of $s_j$ in the field $\mathbb{Z}_2$. Any `$a$' satisfies $a \in \{0,1\}$. By considering bit-by-bit \textit{XOR} operator, $\oplus$, the `$a$' coefficients are utilized for a sequence $\{m_{1,j}, m_{2,j},...\}$ by a relation given as $m_{k,j} = 2a_{1,j}m_{{k-1},j} \oplus 2^2a_{2,j}m_{{k-2},j} \oplus ... \oplus 2^{{s_j}-1}a_{{s_j}-1,j}m_{k-{s_j}+1,j} \oplus 2^{{s_j}}m_{k-{s_j},j} \oplus m_{k-{s_j},j}$. The $m$ values can be arbitrarily chosen provided that $1 \leq k \leq s_j$, and $m_{j,k} \in \{2n+1: n \in \mathbb{Z}_0^{+} \} $ and $m_{j,k} < 2^k$. With a denote of direction numbers, $\{v_{1,j}, v_{2,j}, ... \}$, where any $v_{j,k} = \frac{m_{k,j}}{2^k}$, $j^{th}$ component of the $i^{th}$ point in a Sobol sequence is presented: $x_{i,j} = b_1v_{1,j} \oplus b_2v_{2,j} \oplus...$, where any b is the right-most bits (i.e., least-significant ones) of the $i$ sub-index in binary form.
}

\bibliographystyle{IEEEtran}
\bibliography{bibliography,Hassan}

\begin{IEEEbiography}[{\includegraphics[width=1in,height=1.25in,clip,keepaspectratio]{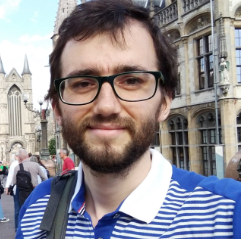}}]{Sercan Aygun}
(S’09-M’22) received a B.Sc. degree in Electrical \& Electronics Engineering and a double major in Computer Engineering from Eskisehir Osmangazi University, Turkey, in 2013. He completed his M.Sc. degree in Electronics Engineering from Istanbul Technical University in 2015 and a second M.Sc. degree in Computer Engineering from Anadolu University in 2016. Dr. Aygun received his Ph.D. in Electronics Engineering from Istanbul Technical University in 2022. Dr. Aygun’s Ph.D. work has appeared in several Ph.D. Forums of top-tier conferences, such as DAC, DATE, and ESWEEK. He received the Best Scientific Research Award of the ACM SIGBED Student Research Competition (SRC) ESWEEK 2022 and the Best Paper Award at GLSVLSI'23. Dr. Aygun's Ph.D. work was recognized with the Best Scientific Application Ph.D. Award by the Turkish Electronic Manufacturers Association. He is currently a postdoctoral researcher at the University of Louisiana at Lafayette, USA. He works on emerging computing technologies, including stochastic and hyperdimensional computing in computer vision and machine learning.
\end{IEEEbiography}

\vspace{-10pt}

\begin{IEEEbiography}
[{\includegraphics[width=1in,height=1.25in,clip,keepaspectratio]{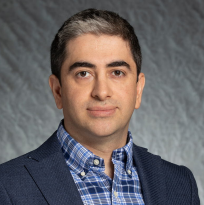}}]{M. Hassan Najafi}
(S’15-M’18-SM'23) received the B.Sc. degree in Computer Engineering from the University of Isfahan, Iran, the M.Sc. degree in Computer Architecture from the University of Tehran, Iran, and the Ph.D. degree in Electrical Engineering from the University of Minnesota, Twin Cities, USA, in 2011, 2014, and 2018, respectively. He is currently an Assistant Professor with the School of Computing and Informatics, University of Louisiana, LA, USA. His research interests include stochastic and approximate computing, unary processing, in-memory computing, and hyperdimensional computing. He has authored/co-authored more than 75 peer-reviewed papers and has been granted 5 U.S. patents with more pending. In recognition of his research, he received the 2018 EDAA Outstanding Dissertation Award, the Doctoral Dissertation Fellowship from the University of Minnesota, and the Best Paper Award at the ICCD’17 and GLSVLSI'23. Dr. Najafi has been an editor for the IEEE Journal on Emerging and Selected Topics in Circuits and Systems.
\end{IEEEbiography}

\vfill

\end{document}